\documentclass[conference]{IEEEtran}
\IEEEoverridecommandlockouts

\usepackage{algorithm}
\usepackage{algorithmic}
\usepackage{graphicx}
\usepackage{amsmath}
\usepackage{amsfonts}
\usepackage{amssymb}
\usepackage{booktabs}
\usepackage{multirow}
\usepackage{xcolor}
\usepackage{pifont}
\usepackage{url}
\usepackage{subcaption}

\newcommand{\cmark}{\ding{51}}%
\newcommand{\xmark}{\ding{55}}%

\begin{document}
\bstctlcite{IEEEexample:BSTcontrol}

\title{CLOAK: Contrastive Guidance for Latent Diffusion-Based Data Obfuscation}

\author{\IEEEauthorblockN{Xin Yang}
\IEEEauthorblockA{
\textit{University of Alberta}\\
Edmonton, AB, Canada \\
xin.yang@ualberta.ca}
\and
\IEEEauthorblockN{Omid Ardakanian}
\IEEEauthorblockA{
\textit{University of Alberta}\\
Edmonton, AB, Canada \\
oardakan@ualberta.ca}
}

\maketitle

\begin{abstract}
Data obfuscation is a promising technique for mitigating attribute inference attacks by semi-trusted parties with access to time-series data emitted by sensors. Recent advances leverage conditional generative models together with adversarial training or mutual information-based regularization to balance data privacy and utility. However, these methods often require modifying the downstream task, struggle to achieve a satisfactory privacy-utility trade-off, or are computationally intensive,
making them impractical for deployment on resource-constrained mobile IoT devices. We propose \textsc{Cloak}, a novel data obfuscation framework based on latent diffusion models. In contrast to prior work, we employ contrastive learning to extract disentangled representations, which guide the latent diffusion process to retain useful information while concealing private information. This approach enables users with diverse privacy needs to navigate the privacy-utility trade-off with minimal retraining. Extensive experiments on four public time-series datasets, spanning multiple sensing modalities, and a dataset of facial images demonstrate that \textsc{Cloak} consistently outperforms state-of-the-art obfuscation techniques, reducing utility loss by up to 7.21\% and privacy loss by up to 5.76\%, and is well-suited for deployment in resource-constrained settings.
\end{abstract}

\section{Introduction}

Ubiquitous and personal sensing through the Internet of Things (IoT), mobile and wearable devices, has driven significant advances in mobile health, home automation, and other domains where the continuous collection of sensor data is essential.
For example, smartwatches equipped with Inertial Measurement Units~(IMUs) capture multivariate time-series that may be transmitted to the cloud, where powerful machine learning models infer useful, non-private attributes, such as the user's physical activity. However, sharing raw sensor data with semi-trusted third parties raises serious privacy concerns~\cite{velykoivanenko2021those}. This is because although these parties are trusted to faithfully perform desired inferences on the data, they may also carry out unwanted inferences to extract and monetize \emph{private attributes} embedded in this data, without the user's knowledge or consent. For instance, motion data collected for fitness tracking could inadvertently reveal sensitive demographic traits or health conditions~\cite{malekzadeh2019mobile}. These \textit{attribute inference attacks} threaten user privacy and pose a barrier to the broader adoption of ubiquitous and personal sensing technologies.

Protecting private attributes in time-series data emitted by sensors presents unique challenges that render many privacy-preserving techniques inadequate. First, this type of data may contain information about private and non-private (i.e. public) attributes that are not directly recorded by the sensors (e.g. motion data captured by an IMU usually contains some information about an individual's age or health condition). While such latent attributes can be uncovered through deep learning, they cannot be easily redacted or perturbed because they are not explicitly represented in the raw data. This problem is compounded by the fact that in a lower-dimensional space, features relevant to private and public attributes often overlap---a phenomenon known as \emph{entanglement}. 
Thus, redacting or perturbing these features to protect privacy can severely degrade data utility, leading to an inevitable trade-off between utility and privacy. 
Second, Differential Privacy~(DP), which is widely regarded as a gold standard for verifiable privacy, provides guarantees that are not well aligned with the objective in our setting. DP mechanisms are primarily designed to ensure that the inclusion or exclusion of a single individual does not significantly affect the output distribution, e.g. when releasing group statistics or training a model. This protects against a broad class of attacks, including membership inference, and can indirectly limit attribute inference in some cases. However, in our setting, the adversary has access to sensor data from a specific individual, and the goal is to prevent inference of a designated private attribute while preserving information about a correlated public attribute. %
Even in the local setting~\cite{kalupahana2023serandip}, DP mechanisms are not designed to prevent inference of latent private attributes, i.e. private attributes that are not directly represented in an individual's reported data.
Moreover, DP guarantees are attribute-agnostic: the injected noise is calibrated to bound worst-case sensitivity and therefore suppresses information indiscriminately. Thus, the amount of noise sufficient to prevent inference of the private attribute could substantially degrade the inference accuracy for the public attribute, when the private and public attributes are correlated, which is typical in sensor datasets.
Applying DP noise to features that correlate with private attributes requires learning a fully disentangled representation---an objective that has proven difficult to achieve in practice~\cite{weggenmann2022dp}.
Third, techniques based on cryptographic primitives, such as secure multi-party computation or homomorphic encryption~\cite{choi2024blind}, are too computationally intensive for deployment on resource-constrained devices.

\begin{figure*}[t]
\centering
\includegraphics[width=\linewidth]{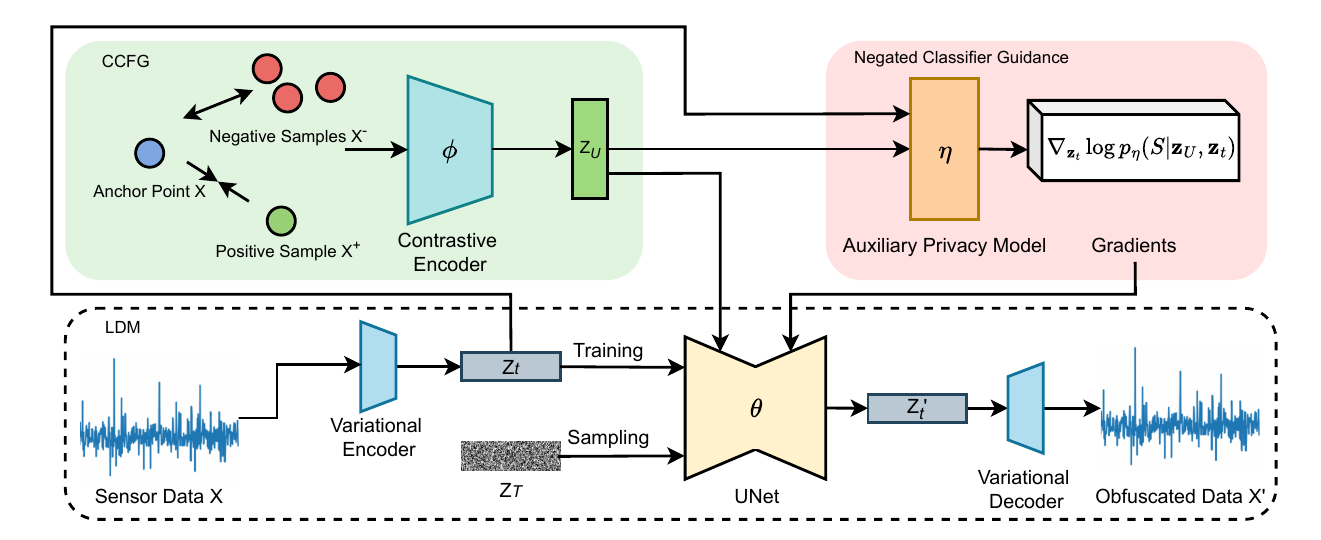}
\caption{Overview of \textsc{Cloak}'s architecture.
}
\label{fig:architecture}
\end{figure*}

Generative models, particularly Generative Adversarial Networks (GANs), ushered in a new paradigm of data obfuscation.
GAN-based obfuscation techniques suppress information about private attributes and preserve information relevant to public attributes in the data~\cite{malekzadeh2019mobile,hajihassnai2021obscurenet}. 
The obfuscated data remains in the same space as the raw sensor data, allowing it to be seamlessly used by downstream applications.
However, GAN-based obfuscation methods suffer from well-known problems. The adversarial min-max game between the generator and discriminator is notoriously unstable and difficult to train. 
More critically, the definition of private attributes shapes the discriminator's objective, which in turn influences the generator during training. Consequently, any change in the private attribute definition or the user's preferred privacy-utility trade-off requires costly retraining of the entire model. These issues make GAN-based approaches poorly suited for accommodating diverse privacy needs or adapting to individual user preferences.
Recent work by Yang \textit{et~al.}~\cite{yang2025privdiffuser} introduced PrivDiffuser, a diffusion model-based obfuscation framework that achieves a strong privacy-utility trade-off, outperforming GAN-based models. However, its obfuscation process is computationally intensive, making it less suitable for deployment on resource-constrained devices. Moreover, PrivDiffuser relies on regularization based on the estimated mutual information to learn disentangled representations of public and private attributes, which is only partially successful in practice.

We propose a novel sensor data obfuscation framework that leverages the power of diffusion models while addressing the limitations of prior work. \textsc{Cloak} utilizes a latent diffusion model to perform obfuscation in a lower-dimensional space, and a variational autoencoder (VAE) to map between the latent and original input spaces. This significantly reduces the computational overhead associated with diffusion models, making \textsc{Cloak} well-suited for deployment on mobile IoT platforms. We introduce a novel conditioning method, named Contrastive Classifier-Free Guidance (CCFG), to guide the latent diffusion process with disentangled public attribute representations, achieving a superior privacy-utility trade-off.
Additionally, we incorporate Negated Classifier Guidance to condition the diffusion model on private attributes, enabling effective removal of information about these attributes at sampling time. Together, these guidance techniques allow \textsc{Cloak} to support diverse user privacy preferences with minimal retraining effort (see Figure~\ref{fig:architecture}).
Our contributions are summarized below:
\begin{itemize}
    \item We propose \textsc{Cloak}, a novel, lightweight obfuscation framework that leverages latent diffusion to mitigate attacks aiming to infer users' private attributes from their time-series data emitted by sensors. 
    
    \item We introduce two novel guidance techniques for \textsc{Cloak}: Contrastive Classifier-Free Guidance to preserve information about the public attribute(s) and Negated Classifier Guidance to suppress information about the private attribute(s), together enabling fine-grained control over the privacy-utility trade-off.

    \item We conduct extensive experiments on five datasets covering multiple sensing modalities. \textsc{Cloak} consistently outperforms state-of-the-art generative obfuscation
    methods on sensor-emitted time-series and images.
    
\end{itemize}

\begin{table*}[t]
\centering
\caption{Comparison between \textsc{Cloak} and prior work on privacy protection for sensor data.}
\label{tab:related_work}
\resizebox{\textwidth}{!}{
\begin{tabular}{lccc}
& Requires no changes to apps & Easy to accommodate diff. privacy needs & Lightweight inference \\ 
\toprule
Privacy-preserving feature extraction~\cite{liu2019privacy,li2021deepobfuscator,li2020tiprdc} 
& \xmark & \xmark & \cmark \\ \hline
GAN-based  obfuscation~\cite{malekzadeh2019mobile, hajihassnai2021obscurenet,raval2019olympus,chen2024mass}
& \cmark & \xmark & \cmark \\ \hline
Diffusion-based  obfuscation w/ MI regularization~\cite{yang2025privdiffuser}
& \cmark & \cmark & \xmark \\ \hline
\textsc{Cloak} (LDM-based obfuscation w/ CCFG)             
& \cmark & \cmark & \cmark \\ 
\bottomrule
\end{tabular}
}
\end{table*}

\section{Related Work}
\subsection{Anonymization and Obfuscation of Sensor Data }
Privacy protection in sensor data can be achieved by learning a low-dimensional representation that contains no information about the private attribute(s) ideally, then sending only this representation to the service provider. 
Early work in this area, such as Privacy Adversarial Network (PAN)~\cite{liu2019privacy} and DeepObfuscator~\cite{li2021deepobfuscator}, focuses on training an encoder for feature extraction in an adversarial fashion, ensuring that a discriminator cannot recognize the private attribute from the learned representation. TIPRDC~\cite{li2020tiprdc} uses mutual information-based regularization instead to minimize the privacy loss.
Since the representations learned by these approaches are not in the original input space, downstream applications must be modified to consume this data, posing a non-trivial challenge for real-world deployment.

To overcome this challenge, conditional generative models have been utilized to generate privacy-preserving synthetic data that has the same dimension as the raw sensor data. 
For example, GANs have been used for editing images and videos to protect sensitive information~\cite{bertran2019adversarially, wu2020privacy,dave2022spact}; however, these methods are not directly applicable to obfuscate time-series as they are specifically designed for visual data based on the higher-dimensional structure.
To obfuscate sensor-generated time-series, Malekzadeh et al.~\cite{malekzadeh2019mobile} proposed training an autoencoder by adding multiple regularization terms to its loss function to reduce privacy loss.
Olympus~\cite{raval2019olympus} and ObscureNet~\cite{hajihassnai2021obscurenet} both used a GAN architecture, training an autoencoder-based obfuscation model, where the autoencoder aims to reconstruct sensor data so as to maintain data utility, and the discriminator acts as an attacker that tries to extract sensitive information from the reconstructed data. 
MaSS~\cite{chen2024mass} transforms sensor data to suppress multiple private attributes for both visual and time-series sensor data by combining information-theoretic measures with adversarial training. 
Yet, MaSS requires fine-tuning the models used in downstream applications, which cannot be done when the service provider is unaware of whether users have obfuscated their data. 
Overall, all these obfuscation models leverage GANs to obscure private attributes embedded in sensor data. However, they lack the flexibility to adapt to the diverse privacy needs of different users, because the training of the discriminator determines the definition of the private attribute, and the weight of the discriminator loss governs the privacy-utility trade-off, both of which are tightly coupled with the training of the generator. Hence, modifying the private attribute requires retraining the entire obfuscation model.

Recently, Yang and Ardakanian~\cite{yang2025privdiffuser} introduced PrivDiffuser, a diffusion-based obfuscation model that achieves a better privacy-utility trade-off than GAN-based obfuscation models. 
To guide the diffusion process, it uses disentangled representations obtained by minimizing mutual information between public and private attribute representations. 
However, as we show later in the paper, this method is less effective at achieving disentanglement compared to \textsc{Cloak}'s CCFG.
Moreover, the sampling process of regular diffusion models is computationally intensive, 
making it difficult to deploy PrivDiffuser on resource-constrained devices.
Table~\ref{tab:related_work} presents a detailed comparison between \textsc{Cloak} and prior work on sensor data obfuscation.

\subsection{Learning Disentangled Representations for Synthetic Data Generation}
A key challenge in sensor data obfuscation is striking a good balance between utility and privacy loss.
This can be addressed by conditioning the generative model with disentangled representations of public and private attributes.
Su \textit{et~al.}~\cite{su2022learning} explored using GAN to disentangle behavior patterns from time-series motion sensor data for activity recognition.
In image editing tasks~\cite{deng2020disentangled,ren2022learning}, 
disentangled representation learning based on contrastive learning and InfoNCE~\cite{oord2018representation} has been successfully integrated with GAN-based generative models.
Wu \textit{et~al.}~\cite{wu2023uncovering} showed that diffusion models, similar to GANs, can inherently learn disentangled latent spaces in text-to-image generation tasks. This property enables controlled image editing on a wide range of attributes without fine-tuning the entire diffusion model.
Yang \textit{et~al.}~\cite{yang2023disdiff} proposed DisDiff, which disentangles the gradient fields of a pre-trained %
diffusion model by minimizing the mutual information.
NoiseCLR~\cite{dalva2024noiseclr} used unsupervised contrastive learning %
to extract disentangled, interpretable image editing directions in diffusion-based text-to-image generation tasks.
DEADiff~\cite{qi2024deadiff} focused on decoupling the style and semantics using Q-formers in attention layers for image style transfer.
These works explore using disentangled representation learning to factorize the latent space of a pre-trained diffusion model. Although they can be applied to edit certain attributes in images, they require powerful diffusion models pre-trained on images and semantics learned from text prompts, such as Stable Diffusion~\cite{rombach2022high} and CLIP~\cite{radford2021learning}, which are not always available for other modalities, such as time-series sensor data.
Another line of research has shown that the factorized latent space of diffusion models can facilitate disentangled representation learning~\cite{leng2023diffusegae, yang2024diffusion, wu2024factorized}.
However, they %
do not investigate using disentangled representations to train conditional diffusion models for privacy protection.

Armandpour et~al.~\cite{armandpour2023re} found that the entanglement of the main prompt with a negative prompt could drastically decrease the quality of the synthesized image in a text-to-image application. They proposed Perp-Neg to alleviate this entanglement by projecting the main and negative prompts to orthogonal spaces.
Yet, this method has shown difficulty in disentangling time-series sensor data~\cite{yang2025privdiffuser}.
\textsc{Cloak} addresses these gaps %
by introducing a new contrastive classifier-free guidance approach.

\section{Threat Model}
\textsc{Cloak} aims to defend against an honest-but-curious (HBC) adversary that has access to sensor data $\mathbf{x}$ shared by a user, intended for desired inferences on the public attribute(s) $U$. 
The adversary can perform intrusive inferences to uncover the user's private attribute(s) $S$ embedded in the shared data, known as attribute inference attacks.
The definition of public attribute $U$ is assumed to be time-invariant as it is tied to the desired downstream applications (e.g. step counting). The private attributes $S$ are specified by users and may change over time.
For other unspecified attributes $R$, we aim to offer some protection since the user may deem them private in the future.
For example, in motion-based human activity recognition, $\mathbf{x}$ represents motion sensor readings, $U$ is the activity class to be inferred from $\mathbf{x}$, $S$ represents the user-specified sensitive attribute (e.g., age) and $R$ represents other unspecified attributes (e.g., height and weight).
We assume that the adversary neither has any metadata about the user, nor does it know if the shared data has been obfuscated, which is a practical assumption since obfuscation may be applied selectively or probabilistically.
The adversary performs both desired inferences and intrusive inferences on a remote server, where the models can be trained on public datasets.

\textsc{Cloak} aims to protect against attribute inference attacks by transforming raw sensor data $\mathbf{x}$ into $\mathbf{x}'$ on the user's device, then sharing $\mathbf{x}'$ with the HBC service provider.
The obfuscation process $f(\mathbf{x})=\mathbf{x}'$ aims to maintain the desired inference accuracy, i.e., $P(U|\mathbf{x})\approx P(U|\mathbf{x}')$, while reducing the intrusive inference accuracy to near random guessing level, i.e., $P(S|\mathbf{x}')\approx P_\text{guess}$.
Note that \textsc{Cloak} does not require knowledge of the intrusive inference model or the desired inference model used by the HBC service provider.

The inherent trade-off between privacy and utility implies that any transformation $f(\mathbf{x})=\mathbf{x}'$ intended to make $P(S|\mathbf{x}')$ close to random guessing would inevitably reduce utility, i.e. $P(U|\mathbf{x}) > P(U|\mathbf{x}')$, and vice versa.

\section{Methodology}
\subsection{Background on Latent Diffusion Models}
Diffusion models are a class of (probabilistic) generative models that have shown extraordinary success in the image domain, outperforming GAN-based approaches~\cite{dhariwal2021diffusion, ho2020denoising}. 
Diffusion models consist of a forward process that gradually adds Gaussian noise through $T$ timesteps to corrupt the data, and a backward process that reverses this process to generate realistic data from noise. 
Given a data point $\mathbf{x}_0 \sim q(\mathbf{x}_0)$, the forward diffusion process perturbs the data at timestep $t$ using noise sampled following a variance scheduler $\beta_t$, where $0<\beta_1<\ldots<\beta_T<1$.
DDPM~\cite{ho2020denoising} models the diffusion process as Markovian:
\begin{equation}
\label{eq:ddpm_xt}
    q(\mathbf{x}_t|\mathbf{x}_{t-1}) = \mathcal{N}(\mathbf{x}_t; \sqrt{1 - \beta_t} \mathbf{x}_{t-1}, \beta_t \mathbf{I}),
\end{equation}
where $\mathbf{x}_t$ is dependent on $\mathbf{x}_{t-1}$, requiring the sampling process to iterate through all $T$ timesteps.
To simplify the sampling process, one can rewrite~(\ref{eq:ddpm_xt}) by defining $\alpha_t = 1-\beta_t$, and $\bar{\alpha}_t = \prod_{i=0}^t \alpha_i$:
\begin{equation}
    \label{eq:ddpm_x0}
    q(\mathbf{x}_t|\mathbf{x}_0) = \mathcal{N}\big(\mathbf{x}_t; \sqrt{\bar{\alpha}_t} \mathbf{x}_0, (1-\bar{\alpha}_t) \mathbf{I}\big),
\end{equation}
where $\mathbf{x}_t = \sqrt{\bar{\alpha}_t}\mathbf{x}_0 + \sqrt{1-\bar{\alpha}_t}\mathbf{\epsilon}_0$ can be derived by applying the reparameterization trick, $\mathbf{\epsilon} \sim \mathcal{N}(0,\mathbf{I})$.
A neural network $\epsilon_\theta(\mathbf{x}_t, t)$ is trained to approximate the intractable $q(\mathbf{x}_{t-1}|\mathbf{x}_t)$ and predict the noise $\epsilon$ added at each timestep. The loss function for training $\epsilon_\theta(\mathbf{x}_t, t)$ is the mean squared error (MSE) given below:
\begin{equation}
\label{eq:mse}
    \mathcal{L}_\theta = \|\epsilon - \epsilon_\theta(\mathbf{x}_t, t)\|_2^2 = \| \epsilon - \epsilon_\theta(\sqrt{\bar{\alpha}_t}\mathbf{x}_0 + \sqrt{1-\bar{\alpha}_t} \epsilon, t) \|_2^2,
\end{equation}
where $t$ is the timestep, $\epsilon \sim \mathcal{N}(\mathbf{0},\mathbf{I})$.

Latent diffusion models (LDMs)~\cite{rombach2022high} extend diffusion models by performing data generation in the latent space $\mathcal{Z}$, which can be obtained using an autoencoder. 
The low-dimensional latent space allows LDMs to focus on semantic embeddings extracted from the original data and is more efficient to compute. %

We train a VAE, comprised of an encoder $\mathcal{E}$ and a decoder $\mathcal{D}$, to learn the latent space $\mathcal{Z}$. The latent representation learned by the VAE, denoted as $\mathbf{z}$, can be expressed as~\cite{kingma2013auto}:
\begin{equation}
    \mathbf{z} = \mu(\mathbf{x};\mathcal{E}) + \epsilon \odot \exp{(\frac{\sigma(\mathbf{x};\mathcal{E})}{2})}, \quad \epsilon \sim \mathcal{N}(\mathbf{0},\mathbf{I}),
\end{equation}
where $\mu$ and $\sigma$ are the mean and log-covariance of the learned multi-variate Gaussian distribution representing the latent distribution of data $\mathbf{x}$. 
Following existing literature~\cite{rombach2022high}, we only use the deterministic posterior generated by the encoder $\mathcal{E}$ as the latent representation for the LDM, i.e., $\mathbf{z}=\mu(\mathbf{x}; \mathcal{E})$. This is because the forward diffusion process already introduces stochasticity, so sampling from the latent distribution can further introduce noise and degrade the quality of data generated by the LDM.
We train a neural network $\epsilon_\theta(\mathbf{x}_t, t)$ to approximate the intractable $q(\mathbf{x}_{t-1}|\mathbf{x}_t)$ and predict the noise $\epsilon$ added at each timestep.
The LDM's loss function can be written as:
\begin{equation}
\label{eq:mse_ldm}
    \mathcal{L}_\theta = \|\epsilon - \epsilon_\theta(\mathbf{z}_t, t)\|_2^2 = \| \epsilon - \epsilon_\theta(\sqrt{\bar{\alpha}_t}\mathbf{z}_0 + \sqrt{1-\bar{\alpha}_t} \epsilon, t) \|_2^2, %
\end{equation}
where %
$\beta$ is the noise scheduler, and
$\mathbf{z}_t = \mathcal{E}(\mathbf{x}_t)$.
Optimizing~(\ref{eq:mse_ldm}) allows the LDM to output generated data $\mathbf{z}_t'$ in the latent space. The decoder $\mathcal{D}$ then maps $\mathbf{z}_t'$ to the input space, i.e., $\mathbf{x}_t'=\mathcal{D}(\mathbf{z}_t')$.
The VAE is pre-trained on the raw sensor data.

\subsection{Contrastive Classifier-Free Guidance for Positive Conditioning}
Let $\mathbf{x} \sim \mathbf{D}$ be a sensor data (time-series) segment, 
a neural network encoder $\phi: \mathcal{X} \to \mathcal{Z}_U$ maps each input to a latent representation $\mathbf{z}_U= \phi(\mathbf{x})$. 
Our goal is to learn representations that contain maximum information about $U$, while being independent of $S$.
To this end, the encoder $\phi$ is trained using contrastive learning.
We condition the LDM with the contrastively learned $\mathbf{z}_U$ via classifier-free guidance, hence we denote our conditioning method as Contrastive Classifier-free Guidance (CCFG).
We consider $\mathbf{x}$ as the anchor point, 
and define the positive/negative sample $\mathbf{x}^+$/$\mathbf{x}^-$ as any data point in $\mathbf{D}$ that shares the same/different public attribute class as the anchor point $\mathbf{x}$, respectively, i.e., $U_{\mathbf{x}^+} = U_\mathbf{x}$, $U_{\mathbf{x}^-} \neq U_\mathbf{x}$.
Note that for both positive and negative samples, their corresponding private attribute classes may differ from the anchor point's private attribute class.
We use the InfoNCE loss denoted as $\mathcal{L}_{\text{InfoNCE}}(\mathbf{x})$ and defined below:
\begin{equation}
    - \!\log \frac{\exp\left( \text{S}(\mathbf{z}_U, \mathbf{z}_U^+)/\tau \right)}{\!\exp\left( \text{S}(\mathbf{z}_U, \mathbf{z}_U^+)/\tau \right) + \sum_{j=1}^K \!\exp\left( \text{S}(\mathbf{z}_U, {\mathbf{z}_U}_j^-)/\tau \right)},
    \end{equation}
where $\mathbf{z}_U^+$ and ${\mathbf{z}_U}_j^-$ are latent representations extracted from positive samples $\mathbf{x}^+$ and negative samples $\mathbf{x}_j^-$, respectively; $K$ is the number of negative samples used;
$\text{S}(u, v) = \frac{u^\top v}{\|u\| \|v\|}$ is the cosine similarity and $\tau$ is the temperature controlling the similarity measurement sensitivity.

The InfoNCE loss approximates a lower bound on the mutual information (MI) $I(\mathbf{z}_U; U)$~\cite{oord2018representation}. Thus, minimizing it 
encourages the encoder $\phi$ to learn a representation $\mathbf{z}_U$ that encodes discriminative information about the public attribute $U$.
Once the encoder is trained, for every data point, we first extract $\mathbf{z}_U=\phi(\mathbf{x})$, and then apply classifier-free guidance~\cite{ho2021classifier} to condition the LDM towards generating data that maintains information about the public attribute $U$.
Following Ho and Salimans~\cite{ho2021classifier}, we define our noise predictor $\epsilon_\theta$ as:
 \begin{equation}
 \label{eq:score_pub}
          \bar{\epsilon}_\theta(\mathbf{z}_t,t,\mathbf{z}_U) = 
        (1+w_U) \epsilon_\theta(\mathbf{z}_t,t,\mathbf{z}_U) - w_U \epsilon_\theta(\mathbf{z}_t,t),
\end{equation}
where $w_U$ is the weight controlling the strength of conditioning the public attribute $U$, $\epsilon_\theta(\mathbf{z}_t,t,\mathbf{z}_U)$ is the score estimator for the LDM conditioned on the contrastively learned %
$\mathbf{z}_U$, and $\epsilon_\theta(\mathbf{z}_t,t)=\epsilon_\theta(\mathbf{z}_t,t, \varnothing)$ is the score estimator for the unconditioned LDM by setting $\mathbf{z}_U$ to zero. 
CCFG then applies Adaptive Group Normalization (AdaGN)~\cite{dhariwal2021diffusion} to condition the LDM on $\mathbf{z}_U$.

\subsubsection{Implicit Disentanglement via Contrastive Sampling}
We now provide the intuition for why $I(\mathbf{z}_U; S)$ is minimized when maximizing $I(\mathbf{z}_U; U)$. 
Our explanation relies on two practical assumptions: 1) the training dataset contains diverse samples, so that there exist many positive pairs $(\mathbf{x}, \mathbf{x}^+)$ that share the same $U$ but differ in $S$; 2) the public and private attributes are not strongly correlated, allowing for at least partial disentanglement between them.

Our sampling strategy ensures that positive pairs $(\mathbf{x}, \mathbf{x}^+)$ are selected solely based on the shared public attribute $U$, irrespective of their private attribute $S$. Thus, samples with the same $U$ but different $S$ must map to nearby representations, discouraging $\phi$ from encoding information about $S$ in $\mathbf{z}_U$.
As a result, the conditional distribution $p(\mathbf{x}^+ \mid \mathbf{x}, U_{\mathbf{x}^+} = U_\mathbf{x})$ will become marginal over $S$, implying a conditional independence relationship in the latent space,
i.e. $\phi(\mathbf{x}) \perp S \mid U$.
Thus, training $\phi$  minimizes the conditional MI ($\min_\phi I(\mathbf{z}_U; S \mid U)$) implicitly.

If $I(\mathbf{z}_U; S \mid U)$ becomes small (ideally approaching 0), then $\mathbf{z}_U$ contains almost no information about $S$ given $U$. 
Thus, minimizing the InfoNCE loss with our sampling strategy leads to representations $\mathbf{z}_U$ that are (i) highly informative of the public attribute $U$, and (ii) almost independent of the private attribute $S$, thereby encouraging disentanglement in the latent space. 
In addition, we note that training the encoder $\phi$ does not require prior knowledge of the private attribute $S$. Therefore, the encoder will learn to disentangle the public attribute from other attributes. This \emph{white-listing characteristic} allows \textsc{Cloak} to extend the privacy protection to unspecified attributes $R$.

\subsection{Enabling Flexible Privacy Protection via Negated Classifier Guidance}

We condition the LDM-based obfuscation model on the private attribute following the idea of classifier guidance~\cite{dhariwal2021diffusion}. This decouples conditioning of the private attribute from the training process, deferring it to the sampling stage. We note that the LDM-based obfuscation model is still conditioned on the public attribute during training, which is acceptable as the public attribute is assumed to be fixed.

Unlike conventional classifier guidance, which is typically used to guide a diffusion model towards adding information about the condition, we use negated classifier guidance such that information about the conditioned private attribute is excluded in the generated data.
Specifically, we apply negated classifier guidance on top of the conditional diffusion model guided by the public attribute $U$. We denote the negated private attribute as $\bar{S}$, i.e., the complement of $S$, and the classifier trained for conditioning the private attribute as the auxiliary privacy model $\eta$ (shown in Figure~\ref{fig:architecture}).
More specifically, $\bar{S}$ is all private attribute classes in the dataset except the user's actual private attribute class.
Following Dong et al.~\cite{dong2023towards} and Liu et al.~\cite{liu2022compositional}, we can write:
\begin{equation}
    \begin{aligned}
    \label{eq:proportional}
    p_{\theta,\phi,\eta}(\mathbf{z}_t| U, \bar{S}) &\propto 
        p_\theta(\mathbf{z}_t)\frac{p_\phi(U|\mathbf{z}_t)}{p_\eta(S|U, \mathbf{z}_t)}.
    \end{aligned}
\end{equation}
Notice that $U$ and $S$ may not be completely disentangled, so we cannot further simplify $p_\eta(S|U, \mathbf{z}_t)$.
Computing the gradient of the log of~(\ref{eq:proportional}) yields:
\begin{equation}
\label{eq:log_proportional}
    \begin{aligned}
    \nabla_{\mathbf{z}_t} \log\big(p_\theta(\mathbf{z}_t)p_\phi(U|\mathbf{z}_t) p_\eta(\bar{S}| U, \mathbf{z}_t)\big) =
    -\frac{\epsilon_\theta(\mathbf{z}_t)}{\sqrt{1-\bar{a}_t}}%
       \\ + \nabla_{\mathbf{z}_t}\log p_\phi(U|\mathbf{z}_t) - \nabla_{\mathbf{z}_t}\log p_\eta(S|U, \mathbf{z}_t),
\end{aligned} 
\end{equation}
This allows writing $\bar{\epsilon}_\theta(\mathbf{z}_t, t, U, S)$ used for sampling as:
\begin{equation}
\label{eq:sample_pub_priv}
\begin{aligned}
   \bar{\epsilon}_\theta(\mathbf{z}_t,t,U,S) =&
    (1+w_U) \epsilon_\theta(\mathbf{z}_t,t,\mathbf{z}_U)  - w_U \epsilon_\theta(\mathbf{z}_t,t) \\+& w_S \sqrt{1-\bar{a}_t} \nabla_{\mathbf{z}_t}\log p_\eta(S|\mathbf{z}_U, \mathbf{z}_t),
\end{aligned}
\end{equation}
where $w_U$ and $w_S$ are weights that control the strength of the positive and negative conditioning, respectively.
To obtain $p_\eta(S|\mathbf{z}_U, \mathbf{z}_t)$, we concatenate the latent representation $\mathbf{z}_U$ with the input of the first FC layer in $\eta$.
$\eta$ must be trained on noisy inputs to consume perturbed $\mathbf{z}_t$ at the sampling stage. 
To overcome this challenge, %
we further apply universal guidance~\cite{bansal2024universal} to allow training $\eta$ on clean data.
During sampling, classifier guidance will be applied to clean data $\hat{\mathbf{z}}_0$ generated by the diffusion model conditioned on the public attribute, rather than the noisy data $\mathbf{z}_t$. 
We express $\hat{\mathbf{z}}_0$ as:
\begin{equation}
    \begin{aligned}
    \label{eq:predicted_clean_data}
        \hat{\mathbf{z}}_0=&\frac{\mathbf{z}_t-\sqrt{1-\bar{\alpha}_{t}}\bar{\epsilon}_\theta(\mathbf{z}_t, t, \mathbf{z}_U)}{\sqrt{\bar{\alpha}_{t}}} \\
        =& \frac{\mathbf{z}_t-\sqrt{1-\bar{\alpha}_{t}} \big( (1+w_U) \epsilon_\theta(\mathbf{z}_t,t,\mathbf{z}_U) - w_U \epsilon_\theta(\mathbf{z}_t,t) \big)}{\sqrt{\bar{\alpha}_{t}}}.
    \end{aligned}
\end{equation}
We can then update our noise prediction model as:
\begin{equation}
\begin{aligned}
   \bar{\epsilon}_\theta(\mathbf{z}_t,t,U,S) =&
    (1+w_U) \epsilon_\theta(\mathbf{z}_t,t,\mathbf{z}_U) - w_U \epsilon_\theta(\mathbf{z}_t,t) \\ +& w_S \sqrt{1-\bar{a}_t} \nabla_{\mathbf{z}_t}\log p_\eta(S|\mathbf{z}_U, \hat{\mathbf{z}}_0).
\end{aligned}
\end{equation}

\section{Experiments}
\label{sec:experiments}

\subsection{Datasets}
We use four publicly available time-series datasets,  MobiAct~\cite{chatzaki2016human}, MotionSense~\cite{malekzadeh2019mobile}, WiFi-HAR~\cite{baha2020dataset}, and AudioMNIST~\cite{becker2024audiomnist}, covering different sensing modalities (i.e. motion, WiFi channel state information, and audio) for evaluation. 
To ensure fairness and consistency, we use the same data selection strategy and pre-processing steps introduced in prior work on data obfuscation. 
Specifically, we follow~\cite{hajihassnai2021obscurenet,yang2025privdiffuser} for MobiAct, MotionSense, and WiFi-HAR; we follow~\cite{chen2024mass} for AudioMNIST.
In addition to sensor-generated (time-series) data, we demonstrate the efficacy of \textsc{Cloak} for obfuscating facial images on an image dataset. 

\subsubsection{MobiAct~\cite{chatzaki2016human}}
It is a human activity recognition (HAR) dataset collected using motion sensors (accelerometer, gyroscope, and orientation sensors) embedded in a smartphone. 
We use the motion data of 36 participants performing 4 different activities: walking, standing, jogging, and walking upstairs. 
We select the data collected from the 3-axis accelerometer and the 3-axis gyroscope, resulting in 6 channels of sensor readings (time-series).
The data is standardized and segmented using a sliding window of size 128 samples with a stride length of 10 samples.
We define the subject's activity as the public attribute. 
We consider 2 private attributes: their gender and weight, where gender contains 2 classes in this case (male, female), and weight contains 3 classes: $\leq$ 70kg, 70-90kg, and $\geq$ 90kg. We split the data into training and testing sets with a ratio of 8:2.

\subsubsection{MotionSense~\cite{malekzadeh2019mobile}} 
\label{subsubsec:motion}
It is another HAR dataset containing motion data collected by the accelerometer and gyroscope in a smartphone. 
We use the motion data of 24 participants performing 4 different activities: walking upstairs, walking downstairs, walking, and jogging.
We consider the subject's activity as the public attribute.
Each activity is repeated 15 times, where the first 9 repeats are used for training and the remaining 6 are used for testing.
We consider the binary gender attribute as the private attribute.
We aggregate the magnitude of accelerometer/gyroscope readings over the 3 axes of each sensor, respectively. This creates a 2-channel sensor data, which is then segmented using the same sliding window as the MobiAct dataset.

\subsubsection{WiFi-HAR~\cite{baha2020dataset}}
This dataset contains WiFi channel state information---a radio frequency-based wireless sensing modality---which can be used to detect human activity. 
We select the data captured by 3 line-of-sight transceiver pairs, comprising 1 transmitter and 3 receivers, where each transceiver pair generates data of 30 channels. Hence, the captured data contains a total of 90 channels.
We compute the magnitude of channel state information (CSI) for each channel and perform standardization. A sliding window of size 80 samples with a hop size of 40 samples is applied to perform data segmentation.
We use the data collected by 10 subjects performing 4 activities: standing, sitting, lying down, and turning around. Each activity is repeated 20 times.
We consider the subject's activity as the public attribute and their weight as the private attribute, where weight is categorized into two classes ($\geq$ 80kg and $<$ 80kg). The data is randomly shuffled and divided into training/test set with a ratio of 8:2.

\subsubsection{AudioMNIST~\cite{becker2024audiomnist}} 
This audio dataset contains recordings of spoken digits from 0-9 in English.
Recordings from 60 participants are included. We use HuBERT-B~\cite{hsu2021hubert} to convert the audio signals into feature embeddings, then rescale the embeddings so that their L2-norm equals 1, following MaSS~\cite{chen2024mass}.
We consider the public attribute to be the digit spoken. We consider the subject's gender (2 classes) as the private attribute,
and their age (18 classes), accent (16 classes), and user ID (60 classes) as unspecified attributes. The train-test split ratio is 8:2.

\subsubsection{Adience~\cite{eidinger2014age}}
This is a facial image dataset used for benchmarking age and gender classification models. We use the aligned faces, a total of 19975 images of 33 subjects. We convert images to grayscale and resize them to 80$\times$80. We then perform normalization to ensure all pixel values are within the range of 0 to 1. We use a 7:3 ratio to randomly sample the training and testing sets.
We consider the 33-class user ID as the public attribute. We consider two private attributes: the user's gender (binary) and the user's age, which is categorized into 8 groups: 0-3, 4-6, 8-13, 15-20, 22-32, 35-48, 48-58, 60-100.

For the Adience dataset, we use a CNN-based VAE, where the encoder contains three batch-normalized convolutional layers. For the encoder, the number of kernels, kernel size, step size, and padding used for each layer are (128, 5, 5, 0), (256, 3, 1, 1), (512, 3, 1, 1), resp. The outputs of the third convolutional layer are fed to two convolutional layers of (8, 3, 1, 1) to output the mean and log variance of the latent distribution.
This maps the grayscale 80$\times$80 image into latent representations with a dimension of 8$\times$16$\times$16. 
We use 3$\times$3 convolutional kernels in the UNet to replace the 1$\times$3 kernels used for time-series data.
We set $w_U{=}6$, $w_S{=}0.5$ for gender obfuscation and $w_U{=}6$, $w_S{=}4$ for age obfuscation.

\subsection{Evaluation Metrics}

\subsubsection{Privacy Loss}
The adversary performs an attribute inference attack to infer private attributes from the shared data, which might be obfuscated. 
We define the privacy loss $L_S$ as the deviation of the classification accuracy of an intrusive inference model ($M_S$) that predicts the private attributes from the obfuscated data from random guessing:
\begin{equation*}
    \label{eq:privacy_loss}
    L_S=|Acc(M_S)-Acc(\text{guess})|\text{, }~Acc(\text{guess})= 
    1/\text{card}(S),
\end{equation*}
where $Acc(M_S)$ is the classification accuracy of the intrusive inference model on the private attribute $S$, $Acc(\text{guess})$ is the random guessing accuracy (assuming the adversary does not know the true distribution of private attribute classes), and $\text{card}(S)$ is the cardinality of the set of private attribute classes.
We use a CNN that contains four convolutional layers followed by three fully connected layers as the intrusive inference model. 
This model is pre-trained on the raw sensor data to predict the private attributes. 
While an adversary could adopt a different architecture, we use this model as a representative example, as it achieves high accuracy on the raw data.
An ideal data obfuscation model should reduce the average intrusive inference accuracy to near the random guessing level; thus, a lower $L_S$ is better. 
For datasets that contain unspecified attributes $R$, we train intrusive inference models $M_R$ in a similar way.

\subsubsection{Data Utility}
The classification accuracy of a desired inference model $M_U$ that predicts the public attribute from the obfuscated data is used as a measure of utility.
Our desired inference model has the same architecture as the intrusive inference model described above.
We pre-train $M_U$ on raw sensor data.
Ideally, a data obfuscation model should generate obfuscated data that maintains the utility of raw data.

\subsection{Implementation}
We pre-train the VAE used in \textsc{Cloak} to transform the sensor data between its original data space and the latent space. 
The VAE is implemented using a symmetric encoder and decoder using
fully connected (FC) layers, where the encoder comprises 4 FC layers with 2048, 2048, 1024, and 512 neurons, resp. The mean and log variance of the latent representation are generated using FC layers with 60 neurons (i.e., dimension of the latent space).
We implement the diffusion model using the UNet~\cite{ronneberger2015u} architecture and adopt a linear schedule from 0.0001 to 0.02 with T=1000. We modify the 2D 3$\times$3 convolutional kernels used in the UNet to 1$\times$3 to work with 1D latent representations.
We use DDIM~\cite{song2021denoising} as the sampling strategy with 50 sampling steps. 
For all datasets, we implement the contrastive encoder $\phi$ %
using a CNN that contains 2 convolutional layers followed by 3 FC layers; 
We implement the auxiliary privacy model $\eta$ using 5 FC layers. We train $\phi$ on raw sensor data and $\eta$ on the latent representation extracted by the VAE's encoder.
We implement \textsc{Cloak} using PyTorch 2 and train it on a server with Ubuntu 20.04, equipped with an Intel Core i9-9940X CPU, 128 GB RAM, and 1 RTX 2080 Ti GPU.
Our implementation is released on GitHub: {\url{https://github.com/sustainable-computing/CLOAK}}.

\subsection{Training Details and Hyperparameter Selection}
We train the VAE used in \textsc{Cloak} using the AdamW optimizer with a learning rate of 0.001. We set the weight for the KL divergence loss to $1e^{-7}$ for AudioMNIST, and $1e^{-6}$ for all other datasets.
We train both the contrastive encoder $\phi$ and auxiliary privacy model $\eta$ using the Adam optimizer with a learning rate of $2e^{-4}$. 
The UNet is trained with the AdamW optimizer with a cosine annealing scheduler that has an initial learning rate of $2e^{-4}$.
We set the training epochs for MotionSense, MobiAct, WiFi-HAR, and AudioMNIST to 80, 28, 80, and 120, resp. 
\textsc{Cloak} uses two hyperparameters $w_U$ and $w_S$ to navigate the privacy-utility trade-offs. We set $w_U$ and $w_S$ empirically through grid search.
For MotionSense, WiFi-HAR, and AudioMNIST datasets, we set $(w_U,w_S)$ to (4.5, 0.008), (6, 0.005), and (2, 0.07), resp. For gender and weight obfuscation on MobiAct, we set $(w_U,w_S)$ to (4, 0.005) and (4, 0.05), respectively.

\subsection{Baselines}
We consider three generative obfuscation models as baselines and evaluate them using the authors' released code.
\subsubsection{ObscureNet~\cite{hajihassnai2021obscurenet}}
It is a GAN-based obfuscation model that has been shown to outperform prior GAN-based obfuscation models. It uses a conditional VAE (CVAE) to reconstruct sensor data and employs a discriminator to remove information about private attributes from the latent space learned by the CVAE. 
We use the randomized anonymization scheme proposed by the authors, as it aligns with our evaluation metric and is robust to re-identification attacks. 
We compare with ObscureNet on MobiAct, MotionSense, and WiFi-HAR.
Since ObscureNet requires training a dedicated obfuscation model for each public attribute class,
it is costly to train it on AudioMNIST, which contains 10 classes for the public attribute (digit).

\subsubsection{MaSS~\cite{chen2024mass}}
It is an obfuscation model that preserves information about both user-annotated public attributes $U$ and unannotated attributes $R$, while protecting private attributes $S$. It integrates adversarial training, mutual information minimization, and contrastive learning, achieving strong performance over several baselines.
However, MaSS makes two assumptions that limit its practicality: 1) downstream applications have access to the obfuscated data and the correct labels for $U$ and $R$ to fine-tune $M_U$ and $M_R$, whereas $\textsc{Cloak}$ does not require changing these models. 
2) data utility with respect to $R$ should be maintained, whereas \textsc{Cloak} aims to protect $R$ because it may be deemed private in the future.
Hence, the results on unspecified attributes $R$ are not directly comparable between MaSS and \textsc{Cloak}.

MaSS is evaluated using an $M_U$ that is fine-tuned for 100 epochs on the obfuscated training dataset and correct labels for $U$, using the same setting as the initial training of $M_U$.
We compare with MaSS on MotionSense and AudioMNIST, as its implementation is available for these datasets only. 

\subsubsection{PrivDiffuser~\cite{yang2025privdiffuser}}
It is a state-of-the-art diffusion-based obfuscation model. It trains a diffusion model that operates in the original data space and
conditions it on the public attribute using latent representations extracted by a surrogate utility model. Classifier guidance is used to condition the private attribute via an auxiliary privacy classifier. PrivDiffuser disentangles public and private attributes by minimizing the estimated mutual information.
We compare with PrivDiffuser on MobiAct, MotionSense, and WiFi-HAR. PrivDiffuser uses 2D convolutional kernels, which do not work for the 1D AudioMNIST dataset.

\begin{table*}[t]
\centering
\caption{\textsc{Cloak}'s obfuscation performance compared to baselines on MotionSense, MobiAct, WiFi-HAR, and AudioMNIST. Results presented as mean$^{\pm \text{std}}$ over 5 runs. Underscored numbers are the best results with statistical significance (p$<$0.05) according to a two-sided paired t-test for MotionSense and a two-sided independent t-test for other datasets.
}
\label{tab:main}
\resizebox{\textwidth}{!}{
\begin{tabular}{llccccc}
\multirow{2}{*}{Dataset}     & \multirow{2}{*}{Method} & Accuracy / F1 Score (\%) & \multicolumn{4}{c}{Privacy Loss $L_S$ (\% deviation from random guessing)}  \\ \cmidrule(l){3-7}
    & & Public Attribute-$U$ & Private Attribute-$S$ & \multicolumn{3}{c}{Unspecified Attributes-$R$} \\ \toprule
\multirow{6}{*}{MotionSense} &              & Activity ($\uparrow$) & Gender ($\downarrow$) & \multicolumn{3}{c}{-} \\ \cmidrule(l){2-7} 
    & Raw Data     & $97.47^{\pm 0.00}$ / $96.59^{\pm 0.00}$ & $43.52^{\pm 0.00}$ & \multicolumn{3}{c}{-}             \\
    & MaSS         & $79.90^{\pm 2.62}$ / $63.76^{\pm 1.97}$ & $6.88^{\pm 0.22}$ & \multicolumn{3}{c}{-}             \\
    & ObscureNet   & $94.96^{\pm 0.33}$ / $93.05^{\pm 0.56}$ & $3.46^{\pm 0.33}$ & \multicolumn{3}{c}{-}             \\
    & PrivDiffuser & $96.32^{\pm 0.56}$ / $95.69^{\pm 0.57}$ & $0.04^{\pm 0.86}$ & \multicolumn{3}{c}{-}             \\
    & \textsc{Cloak}  & \underline{$97.20^{\pm 0.20}$ / $96.28^{\pm 0.31}$} & $0.02^{\pm 0.56}$ & \multicolumn{3}{c}{-}             \\
    \midrule
\multirow{10}{*}{MobiAct}    &              & Activity ($\uparrow$) & Gender ($\downarrow$) & \multicolumn{3}{c}{Weight ($\downarrow$)} \\ \cmidrule(l){2-7} 
    & Raw Data     &  $98.87^{\pm 0.00}$ / $92.04^{\pm 0.01}$ &  $47.51^{\pm 0.00}$  & \multicolumn{3}{c}{$58.41^{\pm 0.00}$}   \\
    & ObscureNet   &  $97.18^{\pm 0.42}$ / $84.80^{\pm 1.32}$ & $2.18^{\pm 0.42}$ & \multicolumn{3}{c}{$39.06^{\pm 1.17}$}   \\
    & PrivDiffuser &  $97.40^{\pm 0.20}$ / $85.49^{\pm 0.64}$ & $1.43^{\pm 0.53}$ & \multicolumn{3}{c}{$16.11^{\pm 1.11}$}   \\
    & \textsc{Cloak}  & \underline{$98.24^{\pm  0.09}$ / $88.25^{\pm 0.69}$} & \underline{$0.06^{\pm 0.61}$}  & \multicolumn{3}{c}{\underline{$8.58^{\pm 1.25}$}}   \\ 
    \cmidrule(l){2-7} 
    &              & Activity ($\uparrow$) & Weight ($\downarrow$) & \multicolumn{3}{c}{Gender ($\downarrow$)}        \\ \cmidrule(l){2-7} 
    & Raw Data     & $98.87^{\pm 0.00}$ / $92.04^{\pm 0.01}$ & $58.41^{\pm 0.00}$ & \multicolumn{3}{c}{$47.51^{\pm 0.00}$}   \\
    & ObscureNet   & $96.97^{\pm 0.28}$ / $84.33^{\pm 1.15}$ & $4.72^{\pm 0.49}$ & \multicolumn{3}{c}{$10.96^{\pm 0.59}$}   \\
    & PrivDiffuser & $97.03^{\pm 0.39}$ / $84.17^{\pm 1.27}$ &  $2.45^{\pm 1.73}$ & \multicolumn{3}{c}{$9.40^{\pm 1.28}$}   \\
    & \textsc{Cloak}  & \underline{$98.24^{\pm 0.09}$ / $88.16^{\pm 0.76}$} &  \underline{$0.14^{\pm 0.42}$} & \multicolumn{3}{c}{\underline{$0.77^{\pm 2.18}$}}   \\ \midrule
\multirow{5}{*}{WiFi-HAR}    &              & Activity ($\uparrow$) & Weight ($\downarrow$) & \multicolumn{3}{c}{-} \\ \cmidrule(l){2-7} 
    & Raw Data     & $97.83^{\pm 1.10}$ / $97.83^{\pm 1.16}$ & $49.13^{\pm 0.11}$ & \multicolumn{3}{c}{-}             \\
    & ObscureNet   & $86.37^{\pm 1.33}$ / $86.17^{\pm 1.32}$ &$ 1.43^{\pm 0.68}$ & \multicolumn{3}{c}{-}             \\
    & PrivDiffuser & $88.18^{\pm 1.37}$ / $88.12^{\pm 1.41}$ & $0.79^{\pm 2.26}$ & \multicolumn{3}{c}{-}             \\
    & \textsc{Cloak}  & \underline{$95.39^{\pm 1.30}$ / $95.39^{\pm 1.30}$} & \underline{$0.18^{\pm 0.38}$} & \multicolumn{3}{c}{-}             \\ \midrule
\multirow{4}{*}{AudioMNIST}  &              & Digit ($\uparrow$) & Gender ($\downarrow$) & Age ($\downarrow$)   & Accent ($\downarrow$) & User ID ($\downarrow$) \\ \cmidrule(l){2-7} 
    & Raw Data     &    $98.92^{\pm 0.02}$ / $98.92^{\pm 0.02}$ &  $48.12^{\pm 0.02}$  & $89.60^{\pm 0.07}$ & $89.05^{\pm 0.03}$ & $94.99^{\pm 0.10}$     \\
    & MaSS         & $97.72^{\pm 0.26}$ / $97.72^{\pm 0.26}$ & $6.10^{\pm 0.32}$ & - & - & -  \\

    & \textsc{Cloak}  & \underline{$98.81^{\pm 0.37}$ / $98.80^{\pm 0.37}$}  & \underline{$0.34^{\pm 0.35}$} & $3.19^{\pm 0.26}$  & $47.35^{\pm 0.86}$  & $0.51^{\pm 0.04}$       \\ 
    \bottomrule
\end{tabular}
}
\end{table*}

\subsection{Evaluation of Data Obfuscation Performance}
\label{subsec:eval_main}
We evaluate obfuscation models on 4 time-series datasets and report average and standard deviation across 5 runs in Table~\ref{tab:main}. 
Our result shows that sharing raw sensor data leads to high privacy loss on all datasets, underscoring the need for obfuscation.

\subsubsection*{Comparison with MaSS} 
MaSS has the worst privacy loss among the baselines. On MotionSense and AudioMNIST, the privacy loss only decreases by 36.64\% and 42.02\%, respectively, relative to the raw data. 
Despite fine-tuning $M_U$ for MaSS, it still underperforms \textsc{Cloak} by 17.30\% and 1.09\% in terms of utility on MotionSense and AudioMNIST, respectively. 
Without fine-tuning $M_U$, the utility of the obfuscated data is severely degraded on both datasets. In particular, without fine-tuning, it  yields 15.83\% accuracy for activity recognition on MotionSense, and 10.28\% accuracy for inferring the pronounced digit on AudioMNIST, effectively rendering the obfuscated data unusable for downstream applications.
It is worth noting that MaSS does not protect unspecified attributes $R$ (it rather does the opposite), hence the privacy loss is not reported for these attributes on AudioMNIST.
\textsc{Cloak} offers reasonable protection for all unspecified attributes without requiring prior knowledge of $R$. The relatively higher privacy loss for the accent attribute is primarily due to the model misclassifying most samples as the dominant class. 
We note that the average F1 score for intrusive inference on accent is only 5.82\%, indicating that the adversary cannot obtain meaningful information on accent from the data obfuscated by \textsc{Cloak}.

\subsubsection*{Comparison with ObscureNet} 
ObscureNet, which is a GAN-based obfuscation model, achieves a good privacy-utility trade-off on MotionSense, MobiAct, and WiFi-HAR.
However, it consistently underperforms the two diffusion-based models--PrivDiffuser and \textsc{Cloak}--in terms of data utility and privacy loss on all four time-series datasets.
Moreover, in the MobiAct dataset, it cannot successfully protect the unspecified attribute.
When performing gender obfuscation on MobiAct, for the unspecified weight attribute, ObscureNet reduces the privacy loss by 19.35\% compared to the raw data, 
while \textsc{Cloak} reduces the privacy loss by 49.83\%.
Similarly, ObscureNet yields the highest privacy loss on the unspecified gender attribute when performing weight obfuscation on MobiAct.
It is also worth noting that ObscureNet must be retrained to protect different private attributes, whereas the contrastive encoder and the diffusion model in \textsc{Cloak} only need to be trained once and can protect different private attributes by introducing the corresponding auxiliary privacy model during sampling.
Apart from these, ObscureNet has a major drawback: it requires training a dedicated model for each public attribute class, significantly increasing the training cost.

\subsubsection*{Comparison with PrivDiffuser}
\textsc{Cloak} achieves comparable privacy loss to PrivDiffuser for the gender attribute on MotionSense, while delivering significantly higher activity recognition accuracy and F1 score.
On MobiAct and WiFi-HAR, \textsc{Cloak} consistently provides the highest data utility with statistical significance.
Notably, on WiFi-HAR, PrivDiffuser can only achieve $\sim$ 88\% activity recognition accuracy given that identifying human activity from radio frequency signals is inherently more challenging. Yet, \textsc{Cloak} drastically improves the activity recognition accuracy and F1 score by respectively 7.21\% and 7.27\%, on average, over PrivDiffuser.
This is while  \textsc{Cloak} achieves the lowest privacy loss on both datasets, significantly outperforming PrivDiffuser.

Considering the unspecified attribute in MobiAct, \textsc{Cloak} reduces privacy loss by 7.53\% compared to PrivDiffuser when performing gender obfuscation. When obfuscating the weight attribute, \textsc{Cloak} reduces the privacy loss by 8.63\% than PrivDiffuser. This highlights the superior disentanglement achieved by the proposed CCFG compared to the mutual information-based regularization used in PrivDiffuser.
Overall, \textsc{Cloak} consistently achieves the best privacy-utility trade-offs across all time-series datasets.

\begin{figure}[t]
    \centering
    \begin{subfigure}[b]{0.49\linewidth}
         \centering
         \includegraphics[width=\linewidth]{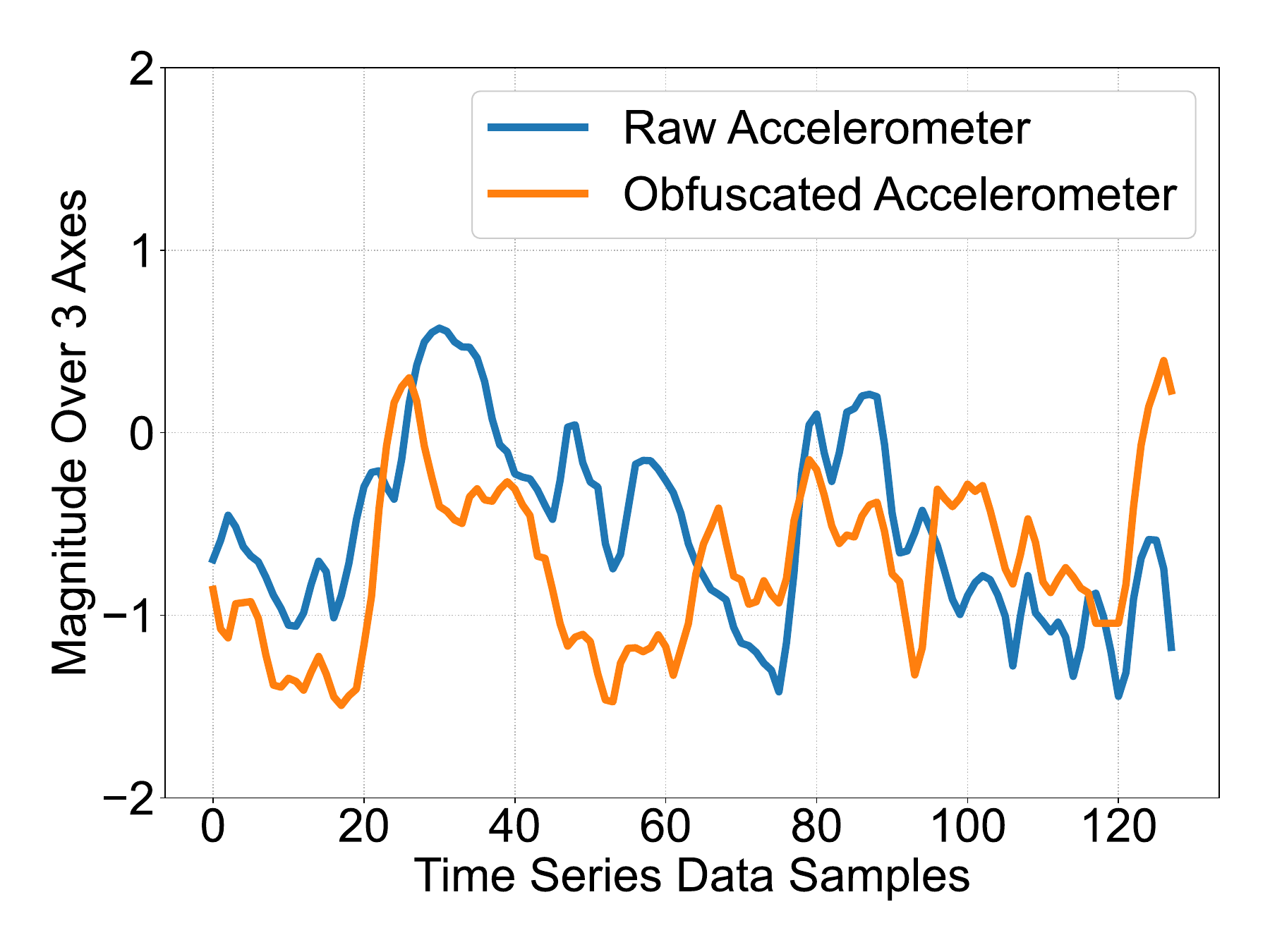}
         \caption{Obfuscating accl. data}
         \label{fig:motion_acc_vis}
     \end{subfigure}
     \hfill
     \begin{subfigure}[b]{0.49\linewidth}
         \centering
         \includegraphics[width=\linewidth]{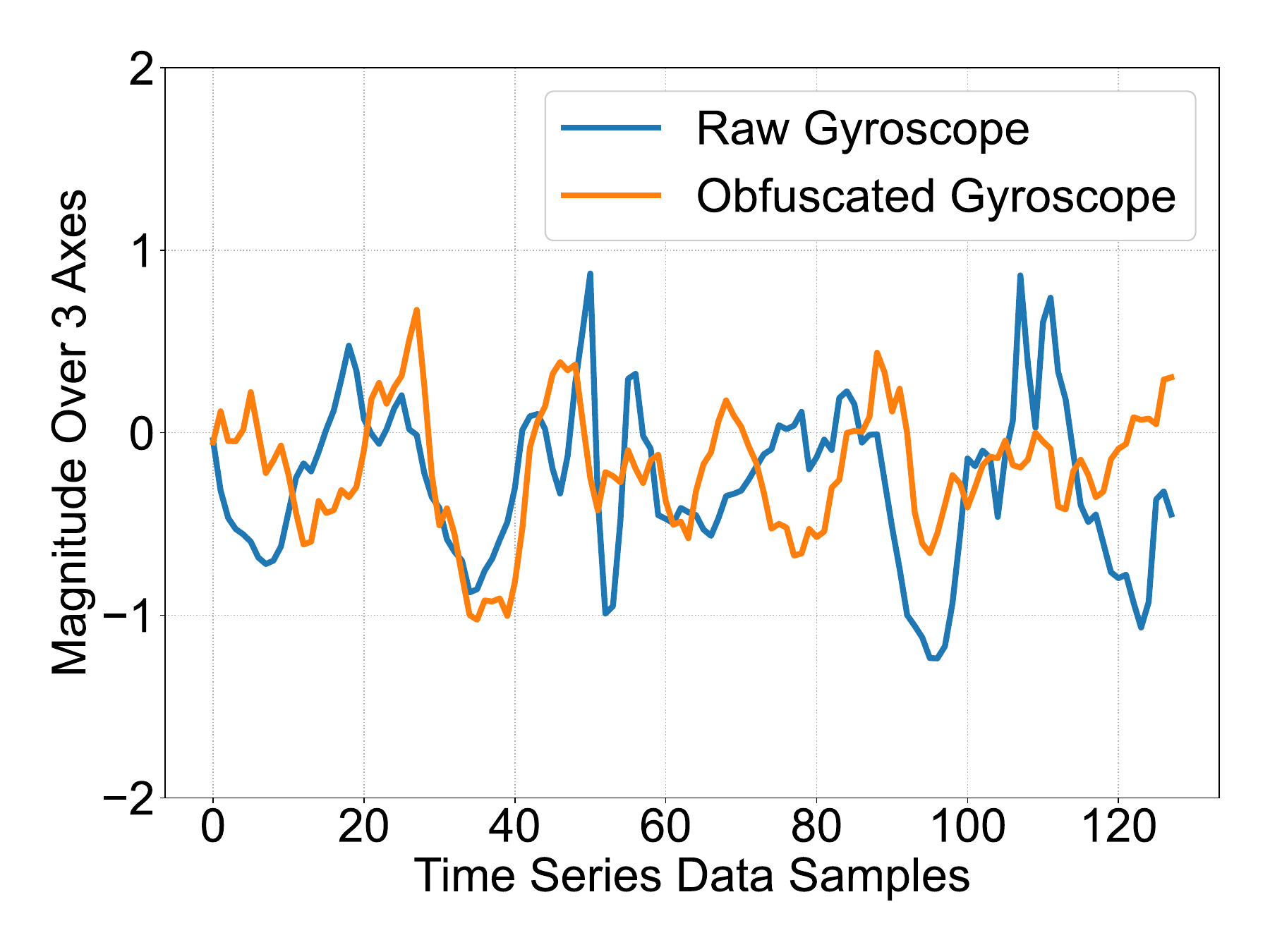}
         \caption{Obfuscating gyroscope data}
         \label{fig:motion_gyro_vis}
     \end{subfigure}
    \caption{Comparison between raw sensor readings and the obfuscated version in MotionSense (public attribute: activity; private attribute: gender)}
    \label{fig:motion_vis}
\end{figure}

\subsubsection*{Visualization of Obfuscated Time-Series Data}
We use the MotionSense dataset as a representative example and compare the original time-series data with those obfuscated by \textsc{Cloak}.  Figure~\ref{fig:motion_acc_vis} shows the original and obfuscated accelerometer data aggregated over the 3 axes, while Figure~\ref{fig:motion_gyro_vis} shows the original and obfuscated gyroscope data aggregated over the axes. 
Each data segment consists of 128 data samples.
The obfuscation process retains the dominant temporal structure of the raw data, including overall trends and the relative spacing between peaks and valleys, thereby preserving the underlying activity patterns.
In contrast, high-frequency components and the amplitude range of the signals are substantially altered. These transformations suppress fine-grained, user-specific motion characteristics that may reveal private attributes such as gender, while retaining coarse-grained structure necessary for desired inference tasks.

\subsection{Protecting Attributes in Images}
We now evaluate \textsc{Cloak}'s obfuscation capability on images from the Adience dataset,
reporting privacy-preserving performance averaged over 5 runs in Table~\ref{tab:adience}.
For comparison, we include the results of MaSS for suppressing the gender and age attributes, respectively, while treating the remaining attributes as unannotated useful attributes.
Because the authors' implementation of MaSS for Adience is unavailable, we were unable to reproduce its results despite our best efforts. Instead, we directly compare with the values reported in~\cite{chen2024mass} and denote them as MaSS* to clarify that the data points and evaluation models used in that study may differ from ours.
Finally, we omit the accuracy for the unspecified attribute because MaSS* is designed to retain, rather than suppress, information about that attribute.

In the case of gender obfuscation, it is evident that \textsc{Cloak} can effectively reduce the privacy loss on the gender attribute to a random guessing level. This comes at the cost of a slight decrease (2.39\%) in data utility.
For the unspecified age attribute, the privacy loss is reduced by 45.24\%. 
This suggests a stronger entanglement between the user ID and the respective age group.
Although there is a gap between the privacy loss on the unspecified age attribute and the random guessing level, the result is encouraging given that this level of protection is provided due to the white-listing characteristic of diffusion-based obfuscation, without requiring any prior knowledge of the user's age.
MaSS* achieves 72.55\% accuracy for user ID classification, which is $\sim$24\% lower than \textsc{Cloak}, meanwhile its privacy loss is $\sim$2\% worse than \textsc{Cloak}.

In the case of age obfuscation, \textsc{Cloak} yields 84.22\% accuracy for user ID classification and 20.40\% privacy loss for the age attribute. We attribute this higher privacy loss to the stronger entanglement between user ID and age, which results in a less favorable trade-off, 
and corroborate this in the next paragraph.
Nevertheless, \textsc{Cloak} provides strong privacy protection for the unspecified gender attribute, yielding a privacy loss of only 2.64\%. 
The results of MaSS* likewise indicate a poorer privacy–utility trade-off when suppressing the age attribute, yielding only 50.05\% desired inference accuracy on the ID attribute, underperforming \textsc{Cloak} by approximately 29\%. Although MaSS* exhibits about 4\% lower privacy loss for the age attribute, \textsc{Cloak} still delivers a substantially better balance between privacy and utility.

\emph{Effect of Public and Private Attribute Entanglement on the Privacy–Utility Trade-off:} 
Table~\ref{tab:adience} shows that data utility decreases substantially, and privacy loss for the private attribute increases substantially when the private attribute is age compared to when it is gender.
To validate that this effect arises from the stronger entanglement between the public attribute (user ID) and age than between the public attribute and gender, we train two MINE~\cite{belghazi2018mutual} models to estimate the MI between user ID and gender, and the MI between user ID and age. We repeat the experiments five times and report the average MI. We find that the average MI between ID and gender is 0.097, whereas the average MI between ID and age is 0.872. This indicates that the correlation between the public attribute and age is indeed much stronger than the correlation with gender, confirming that stronger entanglement can result in a worse privacy–utility trade-off.

\begin{figure*}[t]
\centering
\includegraphics[width=\linewidth]{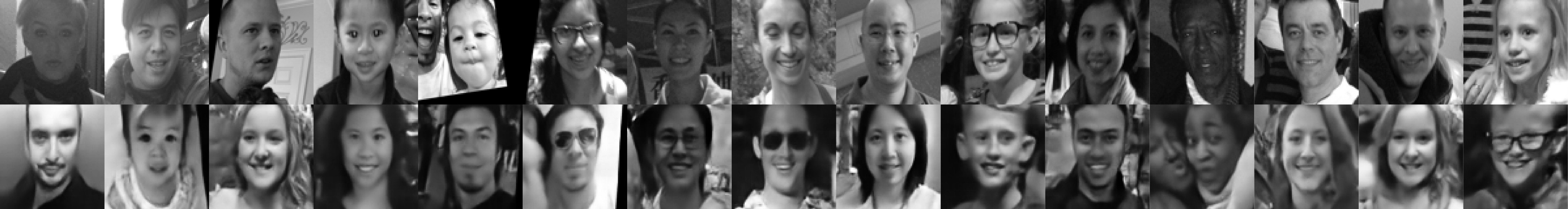}
\caption{Illustration of gender obfuscation on Adience. User ID is the public attribute, age is an unspecified attribute. First row: original images, second row: obfuscated images with the opposite gender detected by the intrusive inference model. Each column is a pair of images before and after obfuscation.}
\label{fig:adience_1pub}
\end{figure*}

\begin{figure*}[t]
\centering
\includegraphics[width=\linewidth]{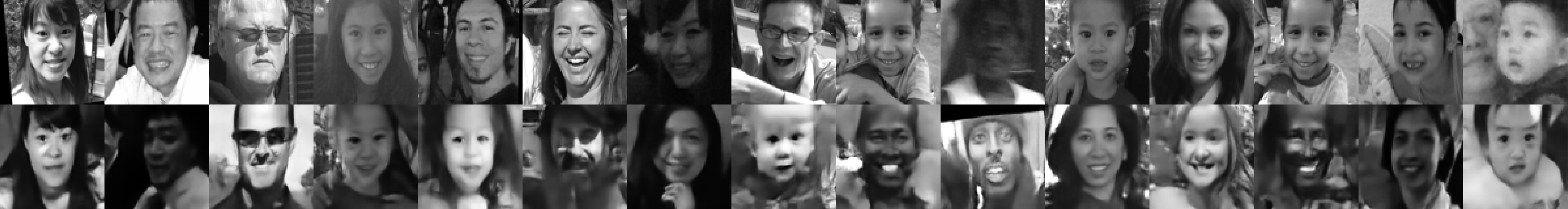}
\caption{Illustration of age obfuscation on Adience. User ID is the public attribute, gender is an unspecified attribute. First row: original images, second row: obfuscated images with a different age group detected by the intrusive inference model. 
Each column is a pair of images before and after obfuscation.}
\label{fig:adience_age}
\end{figure*}

\emph{Visualization of Obfuscated Images:}
In Figure~\ref{fig:adience_1pub} and~\ref{fig:adience_age}, we provide a comparison between images before and after gender and age obfuscation, respectively, where the first row shows the original images, and the second row shows the obfuscated ones. 
For all illustrated images, the desired inference model successfully predicts the correct user ID given the obfuscated images, whereas the intrusive inference model predicts the wrong private attribute class in all cases. 
Looking closely at the obfuscated images, we find that \textsc{Cloak} preserves salient facial characteristics of the user, which are highly associated with their identity.
We also find that the obfuscated images tend to %
borrow features
from the real images of other users in the training set for unspecified attributes, such as hairstyle, age/gender, and background. Thus, unspecified attributes can be protected to a reasonable extent, as long as the training set contains sufficient samples with diverse classes.
We also note that the images generated by \textsc{Cloak} have significantly better quality than the images generated by MaSS (cf. Figure~4 in~\cite{chen2024mass}).
This visualization concurs with the intuition that diffusion-based obfuscation models can protect unspecified attributes by leveraging the inherent stochasticity of sampling in the generative process.

\begin{table}[t]
\centering
\caption{Gender and age obfuscation results on Adience; public attribute is user ID. For MaSS*, we report results from~\cite{chen2024mass} for gender and age suppression, respectively, while the remaining two attributes are unannotated useful attributes.
}

\label{tab:adience}
\begin{tabular}{lccc}
Metric   & Accuracy (\%) & \multicolumn{2}{c}{Privacy Loss $L_S$ (\%)}  \\ \toprule
Attribute  & $U$-ID ($\uparrow$) & $S$-Gender ($\downarrow$) & $R$-Age ($\downarrow$) \\ \midrule
Raw Data & $99.00^{\pm0.10}$ & $49.86^{\pm0.04}$ & $87.38^{\pm0.04}$  \\
MaSS* & $72.55$ & $2.40$ & -  \\

\textsc{Cloak}    & \underline{$96.61^{\pm0.32}$} &  \underline{$0.09^{\pm0.48}$} & $42.14^{\pm0.58}$ \\
\midrule
Attribute  & $U$-ID ($\uparrow$) & $S$-Age ($\downarrow$) & $R$-Gender ($\downarrow$) \\ \midrule
Raw Data & $99.00^{\pm0.10}$ & $87.38^{\pm0.04}$ & $49.86^{\pm0.04}$  \\
MaSS* & $50.05$ & $16.42$ & -  \\
\textsc{Cloak}    & \underline{$84.22^{\pm 1.29}$} &  $20.40^{\pm 1.78}$ & $2.64^{\pm 0.90}$ \\
\bottomrule
\end{tabular}
\end{table}

\subsection{Adjusting Privacy-utility Trade-offs}

\begin{figure}[t]
\centering
\includegraphics[width=0.8\linewidth]{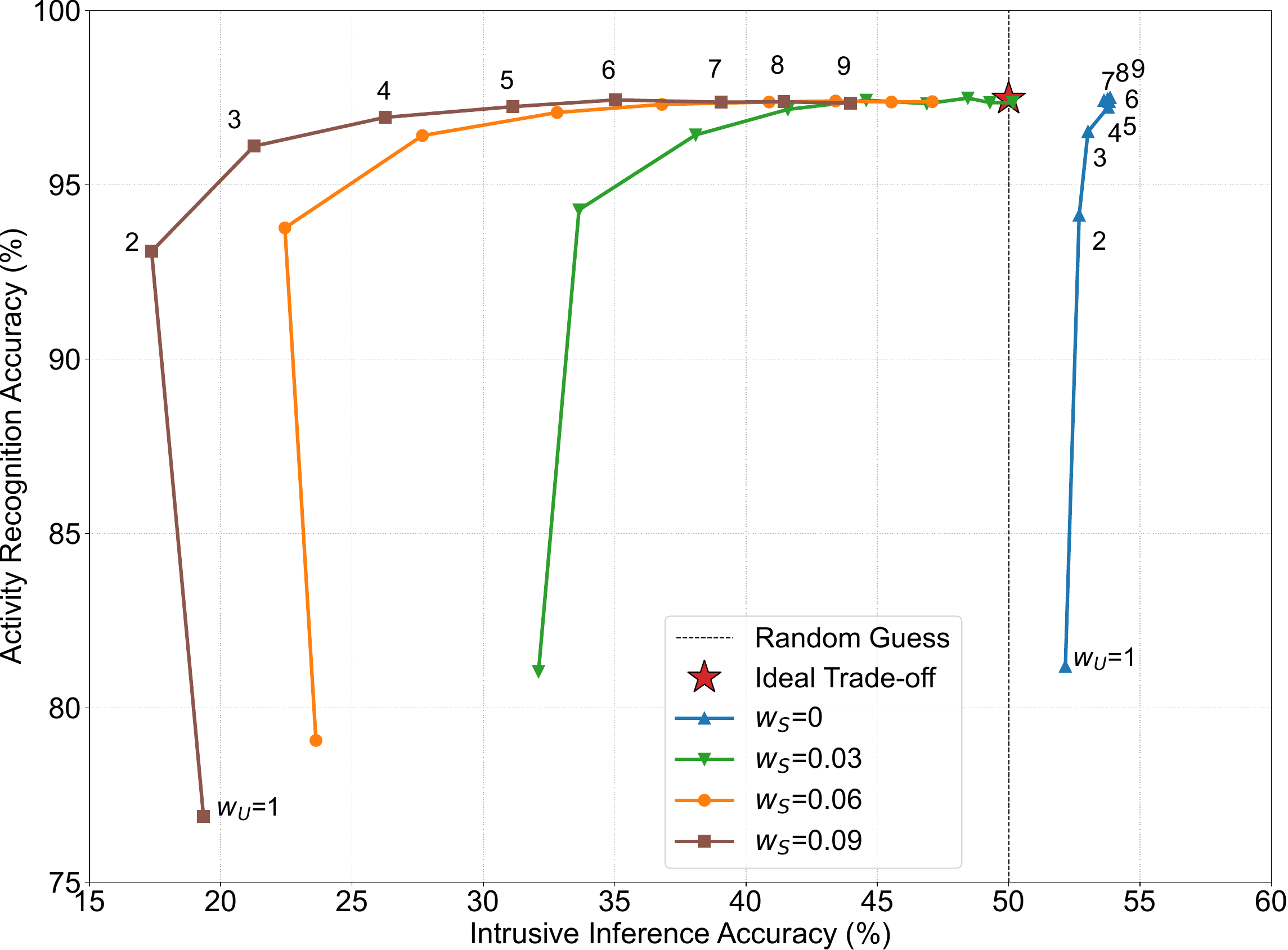}
\caption{\textsc{Cloak}'s privacy-utility trade-off on MotionSense for the gender obfuscation task
}
\label{fig:trade_off}
\end{figure}

\textsc{Cloak} allows users to navigate the privacy-utility trade-off by tuning two hyperparameters, $w_U$ and $w_S$, without retraining.
We demonstrate this for the MotionSense dataset, using activity as the public attribute and gender as the private attribute.
Figure~\ref{fig:trade_off} shows the average desired inference accuracy versus the average intrusive inference accuracy.
The trade-off curves are obtained from a pre-trained \textsc{Cloak} model using a combination of $w_U \in [1,\ldots,9]$ and $w_S \in [0, 0.03, 0.06, 0.09]$. 
The red star denotes the ideal trade-off where the activity recognition accuracy on obfuscated data remains at the same level as raw data and the gender recognition accuracy is at the random guessing level.

For a fixed $w_S$, increasing $w_U$ significantly improves data utility until it reaches $6$. %
However, stronger positive conditioning (higher $w_U$) also increases the intrusive inference accuracy by inevitably including some information about private attributes, as the attributes are not fully disentangled. 
A stronger negative conditioning would be required in this case to balance data utility and privacy. 
In addition, we note that a higher $w_U$ also diminishes the effectiveness of the negative conditioning. Therefore, $w_U$ should be carefully tuned rather than increased indiscriminately.

Fixing $w_U$ and increasing $w_S$ drastically reduces intrusive inference accuracy, highlighting the effectiveness of negative conditioning in removing information about the user's private attribute class.
In particular, when $w_S=0$, 
privacy protection relies solely on \textsc{Cloak}'s white-listing characteristic, resulting in higher intrusive inference accuracy than when $w_S>0$. Moreover, we find that increasing $w_S$ has a limited impact on the activity recognition accuracy, which further confirms that negated classifier guidance can effectively remove information about the private attribute while maintaining data utility.
Overall, a modest level of negative conditioning, e.g., $w_S=0.03$, strikes a favorable balance between utility and privacy on the MotionSense dataset.

\subsection{Quantifying Disentanglement Capability of Contrastive Classifier-Free Guidance}

\begin{figure}[t]
\centering
\includegraphics[width=0.8\linewidth]{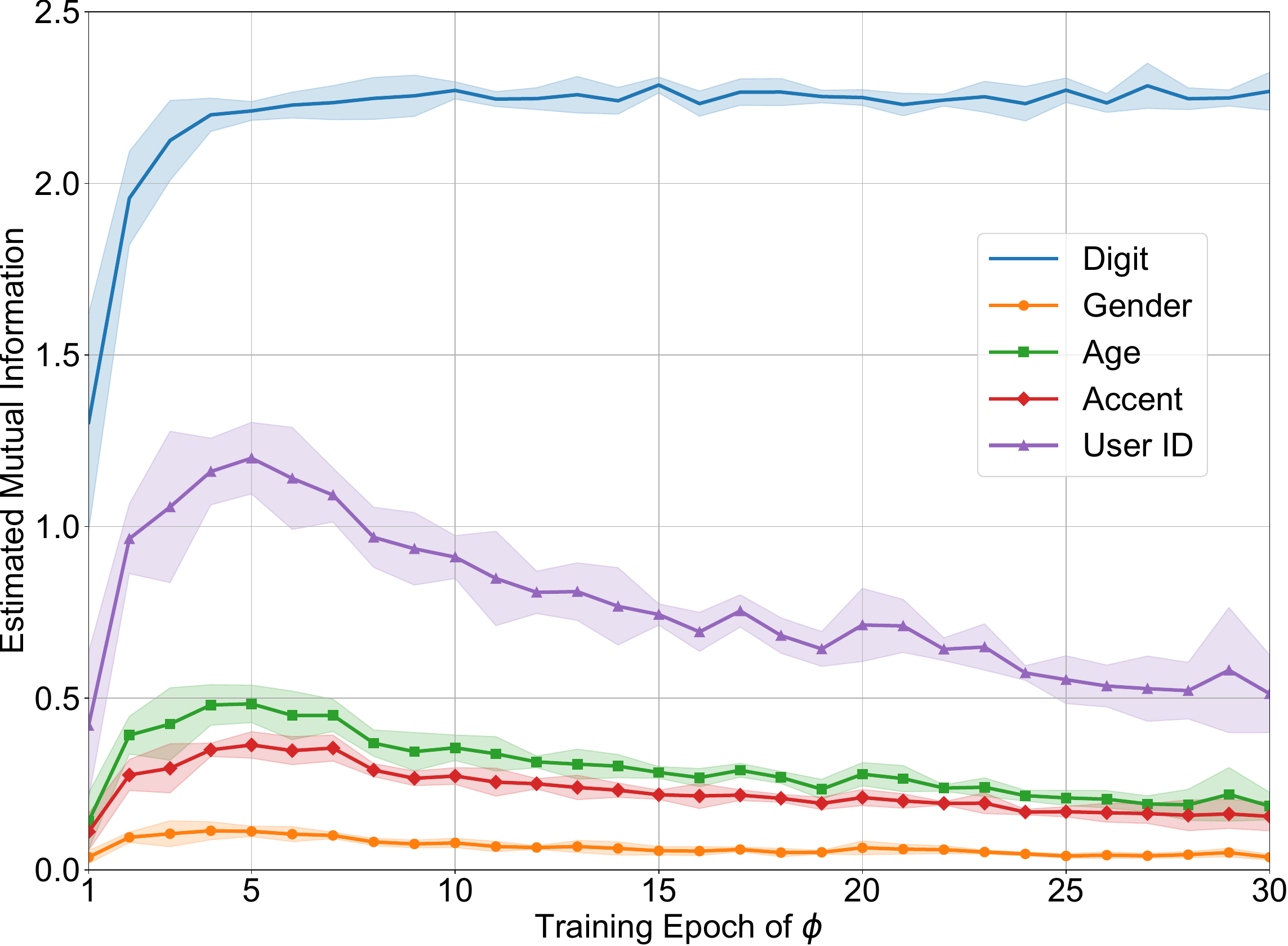}
\caption{Estimated mutual information between contrastively learned $\mathbf{z}_U$ and attributes on AudioMNIST
}
\label{fig:contrastive_mi}
\end{figure}

We corroborate the effectiveness of CCFG in disentangling the public attribute and other attributes on the 5-attribute AudioMNIST dataset, with the spoken digit serving as the public attribute.
The contrastive encoder $\phi$ was trained for 30 epochs, and MI between its learned latent representations ($\mathbf{z}_U$) and every attribute was estimated using the Mutual Information Neural Estimator (MINE)~\cite{belghazi2018mutual}.
Note that CCFG does not require knowledge of the private attribute to perform disentanglement.
Figure~\ref{fig:contrastive_mi} shows the change in MI over 30 training epochs, where the line with markers and color band represent mean$\pm$std
across 5 runs.

It can be seen that the MI between $\mathbf{z}_U$ and the public attribute (digit) consistently increased throughout the training of the contrastive encoder $\phi$, demonstrating CCFG's effectiveness in maintaining utility. 
Conversely, the MI with other attributes, namely gender, age, accent, and user ID, peaked around the 5th epoch, because when $\phi$ is first trained to include information on the public attribute, information on other attributes will also be included due to the entanglement.
However, after the 5th epoch, the MI between $\mathbf{z}_U$ and other attributes significantly decreased by an average of 67.37\% (gender), 61.62\% (age), 57.22\% (accent), and 57.23\% (user ID),
without affecting the MI with the public attribute. 
This confirms that CCFG can disentangle information about the public attribute from all other attributes, thereby improving the privacy-utility trade-off.

\begin{figure}[t]
    \centering
    \includegraphics[width=0.8\linewidth]{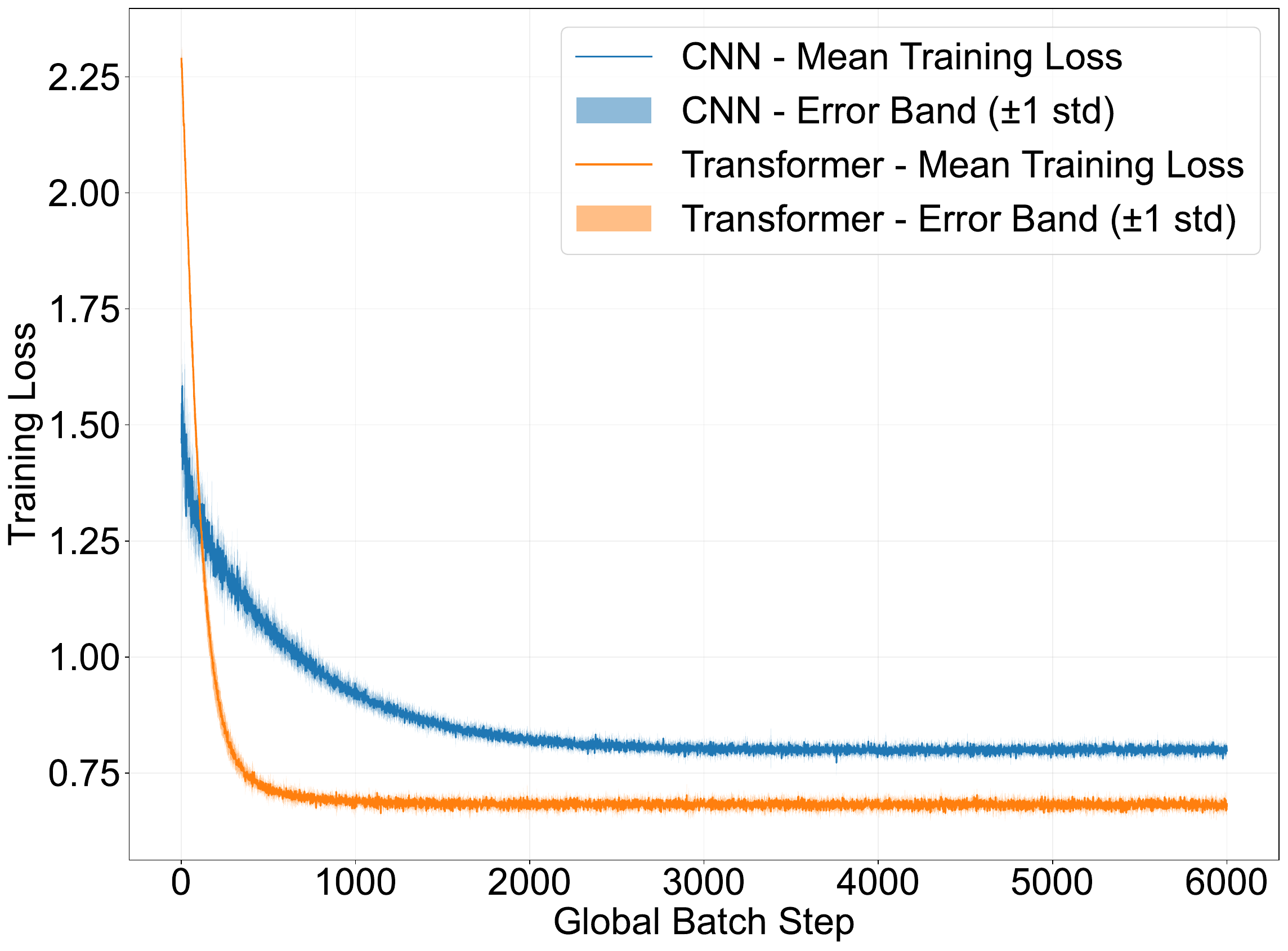}
    \caption{Training loss of $M_R$ when using a CNN or a Transformer-based architecture for the re-identification attack on MotionSense}
    \label{fig:reid_loss}
\end{figure}

\subsection{Defending Against Re-identification Attacks}
In this section, we consider a more powerful adversary that has access to both the obfuscated data and the corresponding ground-truth private attribute labels. 
This adversary performs a re-identification attack~\cite{hajihassnai2021obscurenet} by training a model $M_{R}$ on the obfuscated data to predict the private attribute class.
We use the MotionSense dataset as a case study and reuse the \textsc{Cloak} model presented in Section~\ref{subsec:eval_main} to perform data obfuscation. 
The obfuscated data samples $\mathbf{x}'$ and their associated private attribute labels (gender) are divided into training and test sets following the same ratio as in Section~\ref{subsubsec:motion}. 
$M_{R}$ adopts the same model architecture as the intrusive inference model $M_S$ and is trained using the same pipeline and hyperparameters. 
This architecture is capable of achieving over 93\% accuracy on the raw data, demonstrating sufficient expressive capacity to effectively learn private attribute information from the IMU dataset.
We repeat the experiments for five trials and report the average results.
We find that $M_{R}$ attains an average accuracy of $54.00\%$ in predicting the true gender from the obfuscated data, corresponding to a privacy loss of $4.00\%$, with a standard deviation of $1.26\%$. 
Moreover, we observe that the model has difficulty converging during training, achieving an average training accuracy of only $54.42\%^{\pm1.49\%}$.

In addition, we replace the intrusive inference model $M_{R}$, originally a CNN, with a more powerful Transformer-based classifier which can be used in a re-identification attack. 
This classifier projects the input time-series of shape 2$\times$128 into a 128-dimensional embedding, which is added to the learned positional encoding. The backbone contains 4 pre-norm Transformer blocks with 8-head self-attention, with a head dimension of 16, connected with feed-forward layers of width 256 with GELU activation and residual connections. Global average pooling is then applied to the encoded sequence, followed by a two-layer MLP head with 128 and 32 units, respectively, and a final sigmoid classification layer for binary gender classification.
This Transformer-based model contains 551,233 trainable parameters, nearly 2.8$\times$ more than the CNN-based intrusive inference model.
We train it using the same protocol as the CNN variant and report results averaged over 5 runs.
On the MotionSense dataset, the Transformer-based model achieves an average accuracy of 93.20\% for gender (private attribute) inference on raw data, comparable to the CNN variant.
When we train the Transformer-based intrusive inference model on the data obfuscated by \textsc{Cloak}, the average re-identification accuracy is $57.00\%^{\pm0.02\%}$, only $3\%$ higher than the CNN-based intrusive inference model and near the random guessing level.
We also observe that the Transformer-based intrusive inference model has difficulty converging during training.
We plot the training loss of both CNN and Transformer-based $M_R$ for the first 20 epochs in Figure~\ref{fig:reid_loss}.
Although the Transformer exhibits faster initial convergence than the CNN, both models plateau early and fail to improve further, suggesting that \textsc{Cloak} effectively removes information about the private attribute.

This result indicates that even when an adversary has full access to a sufficient amount of obfuscated data and the corresponding true private attribute labels, training an intrusive inference model to recover the private attribute from obfuscated data remains challenging, as evident from the near-random-guessing performance. This suggests that data obfuscated by \textsc{Cloak} contains minimal information about the private attributes.

\subsection{Conditioning Multiple Public/Private Attributes}
\textsc{Cloak} can be extended by conditioning on multiple public attributes to support more downstream applications or conditioning on multiple private attributes to obscure more than one private attribute.

To support $K$ public attributes $U_1,\ldots, U_K$, for each public attribute $U_i$, 
we utilize a dedicated contrastive encoder $\phi_i$ trained to extract disentangled latent representations $\mathbf{z}_{U_i}$. We condition the diffusion-based obfuscation model with the concatenated latent representations of the $K$ public attributes: $\mathbf{z}_C=\mathbf{z}_{U_1}\oplus \ldots \oplus \mathbf{z}_{U_K}$, then replace $\mathbf{z}_U$ with $\mathbf{z}_C$ in Eq.~(\ref{eq:score_pub}) to train a diffusion model conditioned on the $K$ public attributes.

To protect $K$ private attributes $S_1,\ldots, S_K$, we train $K$ auxiliary classifiers $\eta_1,\ldots, \eta_K$ to predict the corresponding private attribute from the raw sensor data. 
\textsc{Cloak} aims to compute $p_{\theta, \phi, \eta_1,\ldots,\eta_K}(\mathbf{z}_t|U, \bar{S}_1, \ldots, \bar{S}_K)$. 
Assuming that the $K$ private attributes are conditionally independent, we follow the derivation of~(\ref{eq:proportional})~(\ref{eq:log_proportional})~(\ref{eq:sample_pub_priv}) to obtain the updated noise predictor with $K$ private attributes:
\begin{equation}
    \begin{aligned}
        \bar{\epsilon}_\theta(\mathbf{z}_t,t,U,S_1,\ldots,S_K) = (1+w_U) \epsilon_\theta(\mathbf{z}_t,t,U)
       \\ -  w_U \epsilon_\theta(\mathbf{z}_t,t) 
         + \sum_{i=1}^K w_{S_i} \sqrt{1-\bar{a}_t} \nabla_{\mathbf{z}_t}\log p_\eta(S_i|U, \mathbf{z}_t),
    \end{aligned}
\end{equation}
where $w_{S_i}$ is the hyperparameter that controls the strength of negative conditioning for the private attribute $S_i$.
This enables guiding the data obfuscation model during sampling using compositional negated classifier guidance.

\subsubsection*{Experiment results}
We evaluate the effectiveness of conditioning multiple public or private attributes on MobiAct.
For conditioning multiple public attributes, we consider activity and weight as the public attributes, and gender as the private attribute.
For conditioning multiple private attributes, we consider activity to be the public attribute and both gender and weight group as private attributes. The remaining experimental details are the same as we described in the main paper. We report $\text{mean}^{\pm\text{std}}$ computed over 5 runs.

\begin{table}[t]
\centering
\caption{Results of conditioning multiple public attributes on the MobiAct dataset, public attributes: activity (Act.) and weight, private attribute: gender.}
\label{tab:multi_pub}
\begin{tabular}{lccc}
Metric   & \multicolumn{2}{c}{Accuracy (\%)} & $L_S$ (\%)  \\ \toprule
Attribute  & $U_1$-Act. ($\uparrow$) & $U_2$-Weight ($\uparrow$) & $S$-Gender ($\downarrow$) \\ \midrule
Raw Data & $98.87^{\pm0.00}$ & $91.74^{\pm0.00}$ & $47.51^{\pm0.00}$  \\
\textsc{Cloak}    & $94.47^{\pm 1.01}$ &  $80.18^{\pm 2.16}$ & $8.82^{\pm 0.89}$ \\
\bottomrule
\end{tabular}
\end{table}

\begin{table}[t]
\centering
\caption{Results of conditioning multiple private attributes on the MobiAct dataset, public attribute: activity (Act.), private attributes: gender and weight.}
\label{tab:multi_priv}
\begin{tabular}{lccc}
Metric   & Accuracy (\%) & \multicolumn{2}{c}{Privacy Loss $L_S$ (\%)}  \\ \toprule
Attribute  & $U$-Act. ($\uparrow$) & $S_1$-Gender ($\downarrow$) & $S_2$-Weight ($\downarrow$) \\ \midrule
Raw Data & $98.87^{\pm0.00}$ & $47.51^{\pm0.00}$ & $58.41^{\pm0.00}$  \\
\textsc{Cloak}    & $98.25^{\pm 0.12}$ &  $0.79^{\pm 1.90}$ & $0.39^{\pm 0.45}$ \\
\bottomrule
\end{tabular}
\end{table}

We present the results of conditioning \textsc{Cloak} on multiple public attributes in Table~\ref{tab:multi_pub}.
For the public attributes activity and weight, \textsc{Cloak} achieves average desired inference accuracy of 94.47\% and 80.18\%, a decline of 4.40\% and 11.56\% compared to the raw data, respectively.
The average privacy loss of the gender attribute is 8.82\%. 
When \textsc{Cloak} is conditioned on two public attributes, the entanglement between public and private attributes becomes stronger, yielding a worse privacy-utility trade-off compared to when it is conditioned on a single public attribute. Although a larger $w_S$ can further reduce the privacy loss, we notice that the data utility could deteriorate to a greater extent. 
Nevertheless, \textsc{Cloak} still reduces the privacy loss by 38.69\% compared to raw data while offering reasonable data utility. 

Next, we condition \textsc{Cloak} to protect multiple private attributes at the same time, and show the results in Table~\ref{tab:multi_priv}.
\textsc{Cloak} can effectively reduce the privacy loss on both private attributes, gender and weight, to the random guessing level. 
In this case, extending the protection to the weight attribute results in a negligible compromise on data utility. 
Compared to the results in Table~\ref{tab:main}, the average activity recognition accuracy remains at the same level, while the privacy loss for weight is reduced by 8.19\% on average.
This demonstrates that \textsc{Cloak} can be easily extended to protect multiple private attributes.

\begin{table*}[t]
\centering
\caption{Results of conditioning multiple public attributes on the AudioMNIST dataset, public attributes: digit and accent, private attribute: gender.}
\label{tab:audio_2pub}
\begin{tabular}{lccccc}
Metric   & \multicolumn{2}{c}{Accuracy (\%)} & \multicolumn{3}{c}{$L_S$ (\%)}  \\ \toprule
Attribute  & $U_1$-Digit. ($\uparrow$) & $U_2$-Accent ($\uparrow$) & $S$-Gender ($\downarrow$) & $R_1$-Age ($\downarrow$) & $R_2$-User ID ($\downarrow$) \\ \midrule
Raw Data & $98.92^{\pm0.02}$ & $95.30^{\pm0.03}$ & $48.12^{\pm0.02}$ & $89.60^{\pm0.07}$ & $94.99^{\pm0.10}$ \\
\textsc{Cloak}    & $98.09^{\pm 0.23}$ &  $79.16^{\pm 1.03}$ & $5.47^{\pm 3.82}$ & $17.89^{\pm 1.08}$ & $15.59^{\pm 0.51}$ \\
\bottomrule
\end{tabular}
\end{table*}

In addition, to evaluate \textsc{Cloak}'s ability to maintain data utility in more complex settings, we extend the evaluation to conditioning multiple public attributes on the AudioMNIST dataset. Specifically, we consider both spoken digit and speaker accent as public attributes and perform gender obfuscation. We set $(w_U,w_S)$ to (4, 0.3).
The experiments are repeated five times, and the results are reported in Table~\ref{tab:audio_2pub}.
Data obfuscated by \textsc{Cloak} achieves average inference accuracy of 98.09\% and 79.16\% for digit and accent, respectively. 
Compared to the results presented in Table~\ref{tab:main}, adding an additional public attribute (accent) only results in a slight decrease of 0.72\% in digit inference accuracy.
Meanwhile, the privacy loss of the gender attribute is reduced by 42.65\% compared to releasing raw data, effectively preserving data privacy.
Furthermore, due to the white-listing characteristic, \textsc{Cloak} also provides protection for unspecified attributes, i.e., age and user ID, yielding a privacy loss of 17.89\% and 15.59\%, respectively, significantly lower than the privacy loss attained on the raw data.
Overall, these results demonstrate that \textsc{Cloak} remains effective in obfuscating private attributes while maintaining utility for complex downstream tasks, such as simultaneous digit recognition and accent classification in speech data.

\subsection{Evaluation of Computational Overhead}
\label{subsec:overhead}

We compare the computational overhead of \textsc{Cloak} with PrivDiffuser, considering model size, number of parameters, and sampling time per data segment. 
In addition, we deploy \textsc{Cloak} on an Nvidia Jetson to evaluate the real-world obfuscation latency on edge IoT platforms.

We denote the surrogate utility model used in PrivDiffuser and the contrastive encoder $\phi$ used in \textsc{Cloak} as Aux-U, and the auxiliary privacy models used for negative conditioning in both obfuscation models as Aux-S. 
\textsc{Cloak}'s VAE learns latent representations of size 60, and its encoder is not involved in the sampling process. 
In Table~\ref{tab:overhead}, we report the overhead of both obfuscation models measured on MotionSense.
The model size is calculated in FP32 precision. We sample obfuscated data using 50 timesteps per batch, for a total of 50 batches, with a batch size of 128. We report the average GPU execution time for obfuscating one data segment on an Nvidia RTX 2080 Ti.

The Aux-U models used in both obfuscation models share the same architecture, resulting in the same negligible sampling time (0.03 ms).
The Aux-S models used for classifier guidance are also computationally efficient to run and exhibit negligible time overhead. However, the size of Aux-S in \textsc{Cloak} is less than 1\% of the size of Aux-S in PrivDiffuser, as \textsc{Cloak} operates in a latent space that is much smaller than the input space and uses only FC layers rather than the CNN architecture as used in PrivDiffuser.
The size of UNet in \textsc{Cloak} is around $46\%$ smaller than the size of UNet in PrivDiffuser, thanks to the benefit of operating in the latent space. 
Moreover, \textsc{Cloak} uses 1D convolutional kernels in the UNet to accommodate the 1D latent representations, further reducing the overhead of 2D convolutional operations.
As a result, the sampling time of \textsc{Cloak}'s UNet is only 28.85 ms, a speedup of 4.72$\times$ compared to PrivDiffuser.
This indicates that utilizing the LDM architecture for the UNet yields the greatest improvement in the sampling overhead.
Despite the additional model size from the variational decoder, 
\textsc{Cloak} achieves a 1.51$\times$ reduction in the total model size and the number of parameters.
We further evaluate the computational overhead on the Adience dataset for human face obfuscation, where a larger latent representation (dimension 1024) is used to capture richer visual features.
The sampling times for Aux-U, Aux-S, and the decoder are 0.02 ms, 0.00 ms, and 0.04 ms, respectively, indicating negligible overhead.
Despite the latent representation being approximately $17\times$ larger than that used for the MotionSense dataset, the average UNet sampling time is only 79.96 ms. 
Overall, \textsc{Cloak} achieves an average obfuscation latency of 80.02 ms per image, which is faster than the sampling time reported for PrivDiffuser on time-series data.
Given that PrivDiffuser already supports real-time obfuscation on modern edge platforms~\cite{yang2025privdiffuser}, 
\textsc{Cloak}, being more efficient, can be deployed on an even broader range of resource-constrained mobile and IoT devices.

Lastly, we evaluate \textsc{Cloak}'s obfuscation latency on edge IoT platforms by deploying it on an Nvidia Jetson AGX Xavier with 16 GB of memory.
We choose Nvidia Jetson as it is a widely adopted edge AI platform with GPU acceleration. It supports mainstream deep learning frameworks, such as PyTorch, enabling efficient on-device inference and gradient computation. 
Notably, \textsc{Cloak}'s negated classifier guidance requires on-device gradient computation, a capability that is not yet fully supported by many deep learning frameworks designed for smartphones.
We consider the gender-obfuscation task on MotionSense as a case study, demonstrating that \textsc{Cloak} is suitable for deployment on resource-constrained devices. We perform data obfuscation in batches of 128 data segments, while keeping the remaining experimental setups unchanged. We directly run the model trained on a server in FP32 precision without applying any specialized optimization for the Jetson platform. We measure the data-obfuscation time over 20 batches and compute the average latency. On average, obfuscating one batch of data segments takes 26.05 seconds, corresponding to an obfuscation latency of 203.5 ms per data segment. If motion sensors have a sampling rate of 50 Hz and the sliding window used for data segmentation has a stride length of 10 samples, an obfuscation latency of approximately 200 ms would enable real-time obfuscation--closely matching \textsc{Cloak}'s measured latency per data segment. Real-world deployment could further leverage frameworks such as TensorRT and techniques like quantization and lower precision (e.g., FP16) to reduce memory footprint and sampling time.

\begin{table}[t]
\centering
\caption{Overhead comparison between PrivDiffuser (P) and \textsc{Cloak} (C) for gender obfuscation on MotionSense.
}
\label{tab:overhead}
\begin{tabular}{lrrr}
Model  & Size (MB) & \# Params & Samp. Time (ms) \\ \toprule
P: UNet  &  271.05 &  71.05M  &  136.38 \\
P: Aux-U &  130.41 &  34.19M  &    0.03 \\
P: Aux-S &  130.53 &  34.22M  &    0.00 \\ \midrule
P: Total &  531.99 & 139.46M  &  136.41 \\ \midrule
C: UNet  &  146.54 &  38.41M  &   28.85 \\
C: Decoder & 28.14 &  7.38M  &    0.02  \\
C: Aux-U &  130.41 &  34.19M  &   0.03  \\
C: Aux-S &  0.89  &  0.23M  &   0.00  \\ \midrule
C: Total &  305.98 &  80.21M  &   28.90 \\ \bottomrule
\end{tabular}
\end{table}

\section{Limitations and Future Work}
\label{sec:limitation}
While \textsc{Cloak} achieves state-of-the-art obfuscation performance, several limitations remain. First, \textsc{Cloak} requires the user to explicitly specify public and private attribute(s), and its effectiveness depends on the extent to which these attributes can be (at least partially) disentangled. Second, the current evaluation is primarily empirical, complemented by information-theoretic measures of disentanglement. Providing formal privacy guarantees remains challenging, as private attributes are implicitly encoded in sensor data and must be inferred through deep learning models, making it difficult to establish rigorous, model-agnostic bounds. Third, although we demonstrate real-time inference on the Nvidia Jetson platform, representing a meaningful step forward for diffusion-based obfuscation, \textsc{Cloak}'s superior obfuscation performance comes at the cost of higher training and inference overhead compared to GAN-based alternatives. As mobile and IoT devices increasingly incorporate energy-efficient GPU acceleration and deep learning frameworks expand support for on-device gradient computations, we expect \textsc{Cloak} to become deployable across a broader range of platforms including smartphones.

Looking ahead, we plan to investigate the scalability of \textsc{Cloak} in settings involving multiple public and private attributes, as well as its sensitivity to the distribution of data used to train the contrastive encoder. Another direction is to evaluate \textsc{Cloak} on more complex downstream tasks (e.g., automatic speech recognition), where achieving high accuracy is more challenging and attribute entanglement may be more pronounced. Finally, we will consider stronger threat models, including scenarios in which the dataset used to train \textsc{Cloak} is itself sensitive and must be protected.

\section{Conclusion}
\label{sec:conclusion}
We introduce \textsc{Cloak}, a lightweight sensor data obfuscation model based on latent diffusion models.
We propose a contrastive learning-based guidance technique, dubbed CCFG, for conditioning the LDM on public attributes; paired with a negated classifier guidance for conditioning on the private attributes, \textsc{Cloak} offers flexible, fine-grained control over the privacy-utility trade-off without requiring model retraining.
We demonstrate that contrastive learning enables effective information disentanglement in \textsc{Cloak}, allowing it to outperform baselines that rely on mutual information-based regularization or adversarial training.
Specifically, extensive evaluations on four time-series datasets and one image dataset demonstrate that \textsc{Cloak} consistently achieves state-of-the-art privacy-utility trade-offs, improving desired inference accuracy on the public attribute by up to 7.21\%, and reducing intrusive inference accuracy on the private attribute by up to 5.76\%. Moreover, we show that \textsc{Cloak} can be easily extended to protect multiple public or private attributes.
Further, we deploy \textsc{Cloak} on an Nvidia Jetson platform to corroborate that it can perform real-time obfuscation on edge IoT platforms.

\bibliographystyle{IEEEtran}
\bibliography{ref}

\end{document}